\documentclass[11pt]{article}

\usepackage[final]{acl}
\usepackage{times}
\usepackage{latexsym}

\usepackage{amsmath}
\usepackage{booktabs}
\usepackage{multirow}

\usepackage[table]{xcolor}
\newcommand{\graycell}[1]{\cellcolor{gray!15}{#1}}
\usepackage{arydshln}

\usepackage{tcolorbox}
\tcbuselibrary{breakable}

\usepackage[T1]{fontenc}
\usepackage[utf8]{inputenc}

\usepackage{microtype}

\usepackage{inconsolata}

\usepackage{graphicx}

\title{Unlocking Fine-Grained Translation Quality Estimation in LRMs through Mutually Boosting Implicit and Explicit Reasoning}

\author{
    Renfei Dang$^{1}$\text{,} 
    Xinye Wang$^{1}$\text{,} 
    \textbf{Zhejian Lai}$^{1}$\text{,}
    Weilu Xu$^{1}$\text{,}\\
    \textbf{Shimin Tao}$^{2}$\textbf{,}
    \textbf{Daimeng Wei}$^{2}$\textbf{,}
    \textbf{Min Zhang}$^{2}$\textbf{,} 
    \textbf{Shujian Huang}$^{1}$$^\dagger$\\
    $^{1}$ \text{National Key Laboratory for Novel Software Technology, Nanjing University} \\
    $^{2}$ \text{Huawei Translation Services Center, Beijing, China} \\
    \small\texttt{\{dangrf,xinyewang,laizj,weilu.xu\}@smail.nju.edu.cn},
    \small\texttt{huangsj@nju.edu.cn}\\
    \small\texttt{\{taoshimin,weidaimeng,zhangmin186\}@huawei.com}
}

\begin{document}
\maketitle

\renewcommand{\thefootnote}{\fnsymbol{footnote}}
\footnotetext[2]{Corresponding author.}
\renewcommand{\thefootnote}{\arabic{footnote}}

\begin{abstract}
Large Reasoning Models (LRMs) still struggle with fine-grained translation quality estimation (QE), even with long reasoning chains. We argue that LRMs already possess strong multilingual capabilities, while the core challenge stems from the intrinsic difficulty of learning the fine-grained QE task. In this paper, we propose \textbf{RIEQE} (\textbf{R}easoning both \textbf{I}mplicitly and \textbf{E}xplicitly for \textbf{QE}), a simple two-stage training framework that enables the mutually boosting of implicit (layer-wise) and explicit (token-wise) reasoning capabilities. To make implicit reasoning feasible, we first decompose the complex QE task into straightforward subtasks. Based on this, our two-stage approach applies: (1) \textit{NonThinking-SFT}, Supervised Fine-Tuning (SFT) without reasoning chains to directly boost the model's implicit reasoning tendency and capability; and (2) \textit{Thinking-RLVR}, standard Reinforcement Learning with Verifiable Reward (RLVR) to subsequently strengthen explicit reasoning. On the WMT test sets, RIEQE based on Qwen3-4B-Thinking-2507 surpasses all baselines in explicit reasoning performance, while its implicit reasoning capability is also comparable to the best current encoder-based models. We further provide evidence for the mutually boosting between implicit and explicit reasoning, showing how they benefit each other in a bidirectional manner.
Our code is available at \url{https://github.com/NJUNLP/RIEQE}.

\end{abstract}

\section{Introduction}

\begin{figure}[ht]
    \centering
    \includegraphics[width=\linewidth]{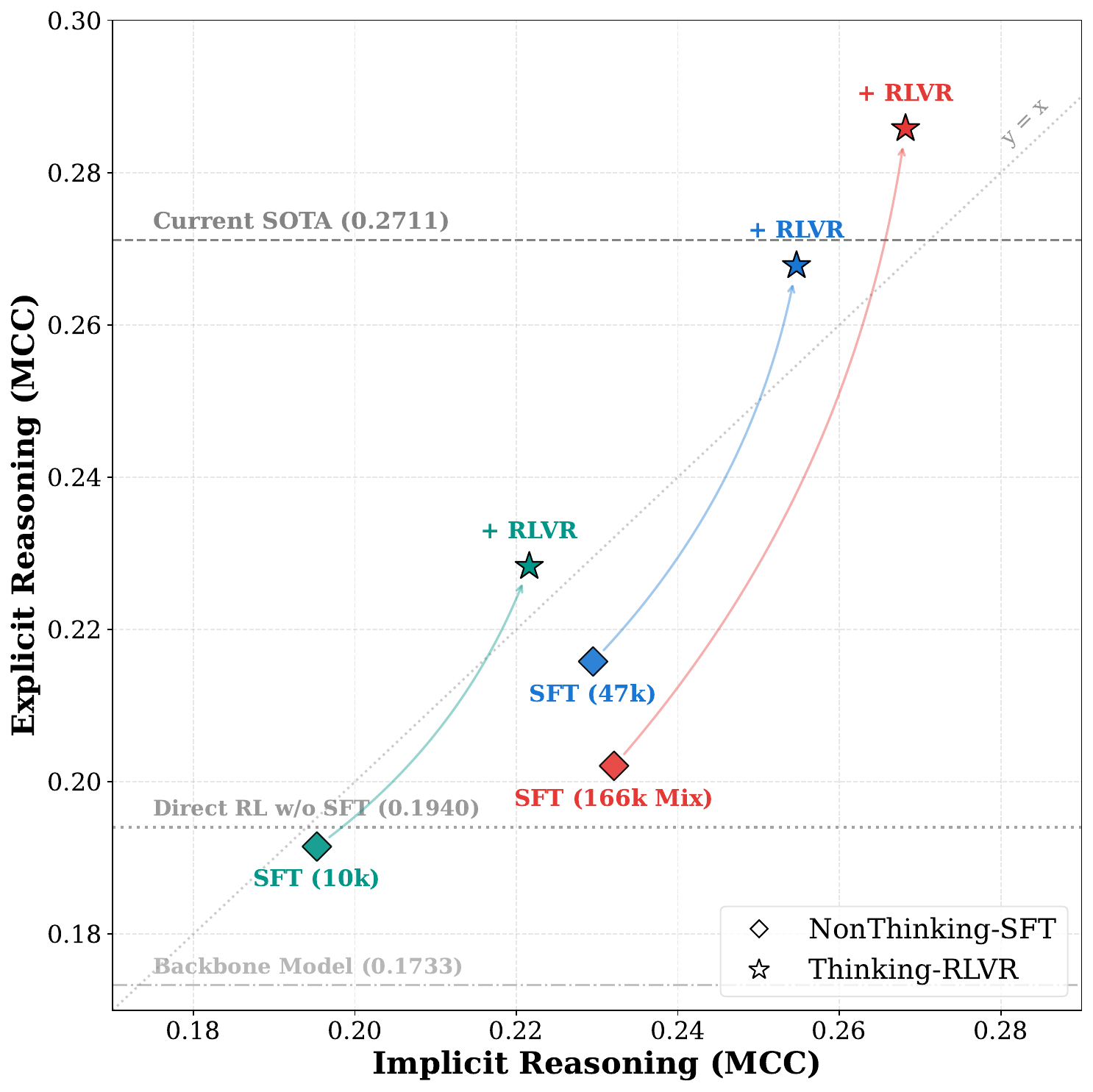}
    \caption{Performance evolution of implicit and explicit reasoning. The horizontal axis represents implicit reasoning performance, while the vertical axis represents explicit reasoning performance. We observe that both SFT and RLVR simultaneously enhance the model's capabilities in both types of reasoning.}
    \label{fig:main_fig_co_eval}
\end{figure}

Translation Quality Estimation (QE) is a technique for evaluating translation quality without relying on reference translations. Previous approaches primarily focus on sentence-level scoring \citep{rei-etal-2020-comet,maheswaran-etal-2025-taser,luo-etal-2025-hw}. 
However, scalar scores (e.g., 70 versus 65) offer neither clear discriminability nor meaningful interpretability.
In contrast, fine-grained quality estimation aims to identify precise error spans (and their severity), providing more usable signals for downstream tasks.

Large Language Models (LLMs) have achieved state-of-the-art (SOTA) translation performance \citep{wmt2025mtfindings}, demonstrating strong multilingual capabilities. Based on LLMs, recent Large Reasoning Models (LRMs) inherit these multilingual strengths while further enabling explicit token-level reasoning for the analysis of translation quality estimation, achieving the best sentence-level QE performance \citep{wmt2025qefindings}. However, they remain unreliable for fine-grained evaluation tasks \citep{NEURIPS2025_5dd3a72b}, falling behind encoder-based models~\citep{wmt2024qefindings,wmt2025qefindings}.

%Results from the Workshop on Machine Translation (WMT) fine-grained quality estimation shared tasks in 2024 \citep{wmt2024qefindings} and 2025 \citep{wmt2025qefindings} indicate that LLMs and LRMs still lag behind encoder-based models on fine-grained QE tasks.

This gap suggests that the potential of LLMs has not been fully elicited in fine-grained discriminative tasks. \citet{huang-etal-2024-lost} showed that LLMs may overlook quality issues in fluent sentences when performing QE. And \citet{tyen-etal-2024-llms} found that when overlooked errors are pointed out by the user, LLMs are able to explain and correct them, suggesting that the models do possess the underlying knowledge but struggle to apply it effectively. Therefore, the key challenge lies not in the lack of multilingual knowledge, but rather in improving the model's ability to locate precise error spans.

Recent studies suggest that the implicit reasoning \citep{ye2025how,lin-etal-2025-implicit} processes of LRMs, namely the layer-wise internal computations that occur without generating explicit thinking tokens, can influence both the process and outcomes of explicit token-level reasoning \citep{dang2026the,chen2025reasoningmodelsdontsay,arcuschin2025chainofthought}. 
This raises the possibility that strengthening implicit reasoning could serve as a foundation for improving explicit reasoning. This strategy is promising for fine-grained QE, where direct RLVR to elicit explicit error-location reasoning is unlikely to succeed when the model struggles to roll out meaningful reasoning trajectories in the first place.
Motivated by these findings, we propose \textbf{RIEQE}, a training paradigm that enhances LRMs' explicit reasoning performance in fine-grained QE tasks by first improving their implicit reasoning ability. Here, implicit reasoning refers to internal hidden-state computations while directly producing final outputs, without introducing latent reasoning tokens. In other words, the LRM directly predicts answers without chain of thought. To make implicit reasoning learning feasible, we decompose the fine-grained QE task into straightforward subtasks. With this decomposition, the learning process proceeds in two stages: (1) \textit{NonThinking-SFT}, which fine-tunes the model using only final outputs without reasoning chains, thereby strengthening its implicit reasoning ability through internal hidden-state computation; and (2) \textit{Thinking-RLVR}, which further encourages the model to spontaneously organize explicit reasoning chains during inference.

% Experimental results demonstrate that our method enables the collaborative co-evolution of explicit and implicit reasoning abilities. 
On the WMT fine-grained QE test sets, RIEQE achieves SOTA performance on three high resource language pairs, Chinese-English (zh-en), English-German (en-de), and English-Russian (en-ru), as well as the low resource language pair, English-Marathi (en-mr). Meanwhile, its implicit reasoning performance without generating thinking tokens remains competitive with the strongest baselines. 

As illustrated in Figure \ref{fig:main_fig_co_eval}, scaling up \textit{NonThinking-SFT} data (i.e., enhancing Implicit Reasoning performance) raises the performance ceiling for subsequent explicit reasoning training; and the subsequent \textit{Thinking-RLVR} stage simultaneously brings significant improvements to both types of reasoning performance.

\section{Background}

\subsection{Fine-Grained Error Detection Tasks}
\label{sec:fineGrainedErrorDetectionTasks}
Following the setup of the WMT QE Shared Task \citep{wmt2024qefindings,wmt2025qefindings}, fine grained error detection is evaluated from two perspectives: word-level and span-level. Word-level QE determines OK/BAD labels for each translation tokens and primarily uses Matthews Correlation Coefficient (MCC) as the evaluation metric. Span-level QE further evaluates the severity of detected errors, which are typically categorized as MINOR and MAJOR\footnote{Sometimes defined as MINOR, MAJOR, and CRITICAL, with most studies merging CRITICAL into MAJOR.}. The main evaluation metric for span-level QE is character level F1 score.

Encoder-based models assign probabilities to each token, allowing predictions to be obtained directly by applying a threshold over token level scores. Auto-regressive LLMs must explicitly generate the predicted error spans and their corresponding severity labels. The generated spans are typically mapped back to the dataset tokenization through a simple rule based alignment procedure.

% \begin{figure*}
%     \centering
%     \includegraphics[width=\linewidth]{pics/decomposition.pdf}
%     \caption{An example of task decomposition. The yellow span ``south turkey'' in STEP3 means MINOR errors recognized by the model, which should be ``southern Turkey''; and the red span ``man'' is a MAJOR error, which should be ``woman''.}
%     \label{fig:decomposition}
% \end{figure*}

\begin{figure*}
    \centering
    \includegraphics[width=\linewidth]{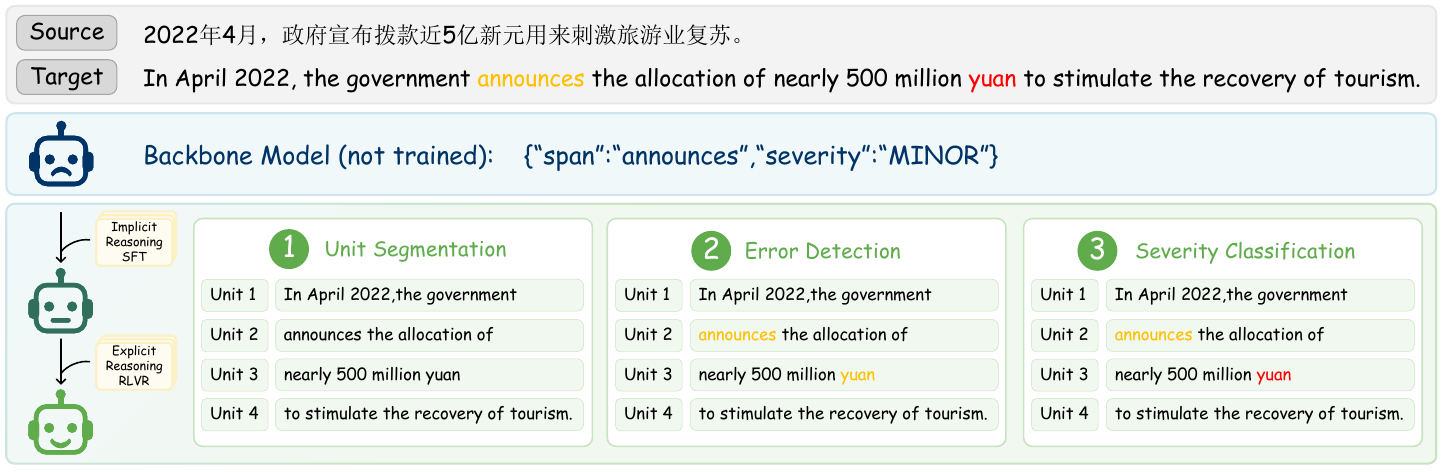}
    \caption{The 2-stage training pipeline of RIEQE and a real example of training effects before and after training. The yellow span indicates a MINOR error, which could better be in past tense; and the red span is a MAJOR error, which should refer to Singapore's currency rather than China's. The latter does not affect fluency, making it difficult to detect before RIEQE training.}
    \label{fig:rieqeexample}
\end{figure*}

\subsection{Challenges for LLMs in Fine-Grained Discriminative Tasks}
Recent works have highlighted potential risks for LLM-as-a-Judge. For instance, LLMs tend to prefer outputs that are closer to their own generation distribution \citep{chen-etal-2025-beyond}. Their evaluations can also exhibit position bias~\citep{wang-etal-2024-large-language-models-fair}, length bias~\citep{zhang2026penalizinglengthuncoveringsystematic}, and language bias~\citep{zhou2026fairnessfluencyinvestigationlanguage}. Some study \citep{huang-etal-2024-lost} further showed that LLMs may be misled by fluency, overlooking the actual quality of a sentence in QE tasks.

Although reasoning models improve QE performance through explicit reasoning \citep{wmt2025qefindings}, the faithfulness of their reasoning remains a concern. \citet{dang2026the} found that implicit layer-level reasoning can influence the reasoning paths and outcomes of explicit reasoning. In some cases, LRMs may even post-hoc rationalize pre-determined answers~\citep{chen2025reasoningmodelsdontsay,arcuschin2025chainofthought}. On the other hand, \citet{tyen-etal-2024-llms} suggested that while LLMs may struggle to identify reasoning errors, they are capable of correcting them when the error locations are provided. These observations offer important insights for LRMs in the QE task, pointing to the potential of leveraging implicit reasoning to improve error localization.

\section{Method}    

In this section, we present the \textbf{RIEQE} framework. We first detail the task decomposition strategy for fine-grained QE to facilitate implicit reasoning learning, and then present our two stage training framework to enhance explicit reasoning by first enhancing implicit discriminative ability.

\subsection{Task Decomposition}
\label{sec:taskDecomposition}
% By decomposing the QE task into more direct and simpler subtasks, the model can more effectively learn implicit discriminative reasoning. 
According to the definition of the fine-grained QE task in \S\ref{sec:fineGrainedErrorDetectionTasks}, the objective is to identify error spans in a translated sentence and assign their severity as MINOR or MAJOR. As illustrated in Figure \ref{fig:rieqeexample}, we decompose the span-level QE task for a translation pair into three sequential subtasks. Specifically, the three subtasks are:
(1) \textbf{UnitSegmentation}: segment the translated sentence into multiple semantically coherent units, making error detection easier by operating on shorter text spans; 
(2) \textbf{ErrorDetection}: examine each semantic unit and determine whether it contains any error span, encouraging the model to thoroughly inspect all sentence content and reduce overlooked errors; 
(3) \textbf{SeverityClassification}: for each detected error span, iteratively predict its severity. 
The prompts for each of these subtasks are presented in Appendix \ref{app:prompts}.

For a given translation pair, if the first step produces $N$ units and there are $E$ errors, we can construct $3+N+E$ training samples in total: one data instance for each of the three subtasks; $N$ instances for unit-level error detection, where the model is asked to find translation error in only one unit; and $E$ instances for severity classification, where the model assigns error severity for only one error span. In the first step, the model may occasionally split a complete error span across two units. In the training set, we merge such cases based on ground truth to avoid misleading the model during learning. This issue cannot be directly avoided in the test set, but it has negligible influence on the results. We provide a more detailed discussion in Appendix \ref{app:errorSpansCrossUnits}.

During inference, the model is also instructed to execute the three subtasks sequentially. For span-level evaluation datasets, the output of the third subtask is directly used as the final prediction. For word-level evaluation datasets, since error severity labels are not required, the output of the second subtask is used as the final prediction instead.

\subsection{Mutually Boosting of Implicit and Explicit Reasoning}
\label{sec:synergy}
RIEQE consists of two stages: \textit{NonThinking-SFT} and lightweight \textit{Thinking-RLVR}.

We first perform SFT on the LRM using decomposed QE data without reasoning chains. 
% Unlike conventional SFT based cold start training,
This \textit{NonThinking-SFT} stage is not merely intended to teach the model the input-output format of the task. Instead, it directly strengthens the model’s tendency and ability to perform implicit discriminative reasoning for fine grained error localization. 

Then, using a dataset constructed from only 1k random sampled translation pairs, we apply \textit{Thinking-RLVR} (hereafter referred to as RLVR) on the second and third subtasks. During this stage, the model learns to spontaneously organize coherent reasoning chains on top of the strengthened implicit reasoning capability. To guide the model toward accurate fine grained error localization and severity prediction, we design task specific rewards for the two subtasks. For subtask 2, ErrorDetection, the reward is defined as the character-level MCC of the translation sentence. For subtask 3, SeverityClassification, the reward is defined as the proportion of correctly predicted error severity labels (MINOR/MAJOR):

\begin{equation}
\text{Reward}_{\text{ErrorDetection}} = \mathrm{MCC}_{\text{char}}
\end{equation}

\begin{equation}
    \text{Reward}_{\text{SeverityClassification}} = \frac{E_{\text{correct}}}{E_{\text{total}}}    
\end{equation}

where $\mathrm{MCC}_{\text{char}}$ denote the character-level MCC score; $E_{\text{correct}}$ is the number of error spans whose severity is correctly predicted, and $E_{\text{total}}$ is the total number of error spans in that sample. The handling of other boundary conditions in the reward is described in Appendix \ref{app:rewardDetails}.

\begin{table*}[ht]
\centering
\footnotesize
\begin{tabular}{c l cc ccc}
\toprule
\multirow{2}{*}{LP} & \multirow{2}{*}{System} 
& \multicolumn{2}{c}{Word-level} 
& \multicolumn{3}{c}{Span-level} \\
\cmidrule(lr){3-4} \cmidrule(lr){5-7}
& 
& MCC & F1 
& F1 & Precision & Recall \\
\midrule

\multirow{11}{*}{zh-en}

% ===== External =====
& \multicolumn{6}{l}{\textit{\textcolor{gray}{Baselines}}} \\
& GPT-5.5 & 0.2520 & 0.2643 & 0.2062 & 0.1208 & 0.7050 \\
& CometKiwi-23 & 0.2690 & - & 0.2720 & - & - \\
% & CometKiwi-23-Ensemble & \underline{0.3020} & - & \textit{0.2880} & - & - \\
& xComet-XXL & 0.2291 & 0.2793 & 0.2196 & 0.1427 & 0.4759 \\
& DCSQE & \textit{0.2812} & 0.2861 & \textit{0.2771} & 0.2219 & 0.3688 \\

\cdashline{2-7}
\noalign{\vskip 0.5ex}

% ===== Qwen inference only =====
& \multicolumn{6}{l}{\textit{\textcolor{gray}{Backbone Model}}} \\
& Qwen3-4B-Thinking-2507 & 0.2195 & 0.2756 & 0.2240 & 0.1551 & 0.4031 \\
& Qwen3-4B-Thinking-2507 (subtask) & 0.2343 & 0.2910 & 0.2384 & 0.1746 & 0.3755 \\

\cdashline{2-7}
% \addlinespace
\noalign{\vskip 0.5ex}

% ===== RIEQE =====
& \multicolumn{6}{l}{\graycell{\textit{\textcolor{gray}{Our Method}}}} \\
& \graycell{\textbf{RIEQE-NonThinking}} 
  & \graycell{\underline{0.2903}} 
  & \graycell{0.3417} 
  & \graycell{\textbf{0.2934}} 
  & \graycell{0.2419} 
  & \graycell{0.3727} \\

& \graycell{\textbf{RIEQE}} 
  & \graycell{\textbf{0.3073}} 
  & \graycell{0.3520} 
  & \graycell{\underline{0.2917}} 
  & \graycell{0.2407} 
  & \graycell{0.3700} \\

\midrule
\multirow{11}{*}{en-de}

% ===== External =====
& \multicolumn{6}{l}{\textit{\textcolor{gray}{Baselines}}} \\
& GPT-5.5 & 0.2507 & 0.2833 & 0.2282 & 0.1425 & 0.5732 \\
& CometKiwi-23 & 0.2150 & - & 0.2350 & - & - \\
% & CometKiwi-23-Ensemble & 0.2460 & - & \textbf{0.2730} & - & - \\
& xComet-XXL & 0.2653 & 0.3183 & 0.2423 & 0.1705 & 0.4181 \\
& DCSQE & \underline{0.2711} & 0.3061 & \underline{0.2589} & 0.2120 & 0.3326 \\

\cdashline{2-7}
\noalign{\vskip 0.5ex}

% ===== Qwen inference only =====
& \multicolumn{6}{l}{\textit{\textcolor{gray}{Backbone Model}}} \\
& Qwen3-4B-Thinking-2507 & 0.1581 & 0.2173 & 0.1873 & 0.1655 & 0.2157 \\
& Qwen3-4B-Thinking-2507 (subtask) & 0.1733 & 0.2226 & 0.1970 & 0.1905 & 0.2040 \\

\cdashline{2-7}
\noalign{\vskip 0.5ex}

% ===== RIEQE =====
& \multicolumn{6}{l}{\graycell{\textit{\textcolor{gray}{Our Method}}}} \\

& \graycell{\textbf{RIEQE-NonThinking}} 
  & \graycell{\textit{0.2682}} 
  & \graycell{0.3107} 
  & \graycell{\textit{0.2500}} 
  & \graycell{0.2568} 
  & \graycell{0.2435} \\

& \graycell{\textbf{RIEQE}} 
  & \graycell{\textbf{0.2858}} 
  & \graycell{0.3263} 
  & \graycell{\textbf{0.2643}} 
  & \graycell{0.2742} 
  & \graycell{0.2550} \\

\bottomrule
\end{tabular}
\caption{Word-level and span-level results on zh–en and en–de fine-grained quality estimation. "LP" stands for "Language Pair". The \textbf{bolded} values represent the best results in this column, the \underline{underlined} values represent the second-best, and the \textit{italicized} values represent the third-best. ``(subtask)'' means a 3-step subtask evaluation process.
}
\label{tab:resultswithwordlevel}
\end{table*}

\begin{table*}[ht]
\centering
\footnotesize
\begin{tabular}{c l cccc}
\toprule
\multirow{2}{*}{LP} & \multirow{2}{*}{System} & \multicolumn{4}{c}{Word-level} \\
\cmidrule(lr){3-6} 
& & MCC & F1 & Precision & Recall \\
\midrule

% ===================== en-ru =====================
\multirow{9}{*}{en-ru}

% External
& \multicolumn{3}{l}{\textit{\textcolor{gray}{Baselines}}} \\
& GPT-5.5 & 0.2038 & 0.2541 & 0.1584 & 0.6418 \\
& xComet-XXL & 0.2630 & 0.3286 & 0.2758 & 0.4067 \\
& DCSQE & \underline{0.3510} & 0.3733 & 0.2984 & 0.4985 \\

\cdashline{2-6}
\noalign{\vskip 0.5ex}

% Base model
& \multicolumn{3}{l}{\textit{\textcolor{gray}{Backbone Model}}} \\
& Qwen3-4B-Thinking-2507 & 0.1829 & 0.2533 & 0.1683 & 0.5126 \\
& Qwen3-4B-Thinking-2507 (subtask) & 0.2508 & 0.3120 & 0.1959 & 0.5045 \\

\cdashline{2-6}
\noalign{\vskip 0.5ex}

% RIEQE
& \multicolumn{5}{l}{\graycell{\textit{\textcolor{gray}{Our Method}}}} \\
& \graycell{\textbf{RIEQE-NonThinking}} & \graycell{\textit{0.3434}} & \graycell{0.3871} & \graycell{0.3125} & \graycell{0.5086} \\
& \graycell{\textbf{RIEQE}} & \graycell{\textbf{0.3583}} & \graycell{0.4092} & \graycell{0.3325} & \graycell{0.5318} \\

\midrule

% ===================== en-mr =====================
\multirow{11}{*}{en-mr}

% External
& \multicolumn{3}{l}{\textit{\textcolor{gray}{Baselines}}} \\
& GPT-5.5 & 0.1855 & 0.2747 & 0.1954 & 0.4627 \\
& CometKiwi-23 & 0.2520 & - & - & - \\
& xComet-XXL & 0.1453 & 0.2512 & 0.1979 & 0.3438 \\
& DCSQE & \textit{0.2806} & 0.3423 & 0.2517 & 0.5347 \\

\cdashline{2-6}
\noalign{\vskip 0.5ex}

% Base model
& \multicolumn{3}{l}{\textit{\textcolor{gray}{Backbone Model}}} \\
& Qwen3-4B-Thinking-2507 & 0.0915 & 0.1871 & 0.2053 & 0.1719 \\
& Qwen3-4B-Thinking-2507 (subtask) & 0.0467 & 0.1202 & 0.1454 & 0.1024 \\

\cdashline{2-6}
\noalign{\vskip 0.5ex}

% RIEQE
& \multicolumn{5}{l}{\graycell{\textit{\textcolor{gray}{Our Method}}}} \\
& \graycell{\textbf{RIEQE-NonThinking}} & \graycell{\textbf{0.2868}} & \graycell{0.3361} & \graycell{0.4126} & \graycell{0.2836} \\
& \graycell{\textbf{RIEQE}} & \graycell{\underline{0.2847}} & \graycell{0.3331} & \graycell{0.4223} & \graycell{0.2750} \\

\bottomrule
\end{tabular}
\caption{Word-level results on en-ru and en–mr fine-grained quality estimation. "LP" stands for "Language Pair". The \textbf{bolded} values represent the best results in this column, the \underline{underlined} values represent the second-best, and the \textit{italicized} values represent the third-best.}
\label{tab:resultswithspanlevel}
\end{table*}

\section{Experiment}
\label{sec:experiments}
\subsection{Experiment Setup}
\paragraph{Datasets.}
We use the training data provided by the WMT QE Shared Task~\citep{blain-etal-2023-findings}. Since WMT may incorporate the previous year’s test data into the following year’s training set, we carefully control the training data by year to avoid test data leakage. For evaluation, we use the high-resource WMT 2023 span-level test sets for en-de and zh-en, as well as the WMT 2022 en-ru word-level test set, and the low-resource WMT 2023 en-mr word-level test set. Many hard-to-reproduce baselines are evaluated on these test sets in their original paper, so we directly utilize their results from the papers. More recent test sets did not release official gold labels. More detailed data quantities and ratios are provided in Appendix \ref{app:datadetails}.

\paragraph{Baselines.}
As baselines for WMT 2024 and 2025 fine-grained QE tasks, we include \textbf{CometKiwi-23}~\citep{rei-etal-2023-scaling} and \textbf{xCOMET}~\citep{guerreiro-etal-2024-xcomet}. 
Results of CometKiwi are from its paper; xCOMET is open-source\footnote{\url{https://huggingface.co/Unbabel/XCOMET-XXL}}, and we report results from our local evaluation. 
Other recent LLM-based methods from WMT shared tasks generally underperform these baselines. 
To the best of our knowledge, the current state-of-the-art fine-grained QE model is \textbf{DCSQE}~\citep{geng-etal-2025-alleviating}, which is also based on an encoder architecture with an XLMR-L~\citep{conneau-etal-2020-unsupervised} backbone and is trained on 500k synthetic QE instances. Its original paper reports results on en-de and zh-en, while the results on en-ru and en-mr are obtained from our local reproduction. We also include GPT-5.5 \citep{openai2026chatgpt} as the strong LRM baseline.

\paragraph{Implementation and evaluation.} The backbone model is \textit{Qwen3-4B-Thinking-2507}~\citep{yang2025qwen3technicalreport}. 
For each \textit{NonThinking-SFT} sample, the input includes an empty reasoning chain, 
%(\verb|<think>\n\n</think>|), 
and the output is ground truth answer. We apply LoRA \citep{hu2022lora} to prevent the collapse of the inherent explicit reasoning capability using LlamaFactory framework \citep{zheng2024llamafactory}.
For \textit{RLVR}, we employ GRPO~\citep{shao2024deepseekmathpushinglimitsmathematical} using the verl framework \citep{verl}. Other training details are in Appendix \ref{app:trainingDetails}. 
During testing, we refer to the setting where the model performs explicit reasoning as \textbf{RIEQE}, and the implicit reasoning setting that skips explicit reasoning as \textbf{RIEQE-NonThinking}. Note that both settings undergo two-stage training, differing only in testing methodology.
We perform decoding five times with majority voting to enhance stability. More evaluation details see Appendix \ref{app:multiple-deocde}.

\subsection{Main Results}

\paragraph{RIEQE achieves SOTA performance on fine-grained QE benchmarks.} As shown in Tables \ref{tab:resultswithwordlevel} and \ref{tab:resultswithspanlevel}, RIEQE significantly outperforms strong baselines across four language pairs at both span-level and word-level. The backbone model performs poorly on fine-grained QE, and adopting the three-step subtask evaluation process brings only limited improvements. In contrast, RIEQE helps the model achieve substantial performance gains and reaches SOTA results on fine-grained QE benchmarks. For comparison, the human annotation consistency reported in the WMT 2025 Span level Error Detection Task shows F1 scores of only 0.12-0.35 on high resource language pairs (e.g., en-ja, en-zh, en-ru, en-it) \citep{wmt2025qefindings} (Detailed in Appendix~\ref{app:humanAnnotatorConsistency}). This suggests that the current performance has already reached a relatively high level, and that the remaining room for improvement may primarily reflect better alignment with individual annotator preferences rather than genuinely more universal error detection capability.

\paragraph{RIEQE achieves higher precision.} Across all language pairs, RIEQE consistently achieves the highest precision among all baselines while maintaining relatively strong recall. This indicates that RIEQE yields more accurate error localization rather than simply exhibiting a model-favored precision-recall trade-off. Specifically, the precision of RIEQE ranges from approximately 0.25 to 0.42 across different test sets, also surpassing the human annotators' consistency in precision score as shown in Appendix~\ref{app:humanAnnotatorConsistency}. The high precision means fewer false positives, which is crucial for real-world QE applications by providing more trustworthy and actionable error detection results.

\paragraph{The implicit reasoning ability of RIEQE is also highly competitive.}
RIEQE-NonThinking remains highly competitive with baselines across all metrics without using any thinking tokens. It surpasses other baselines on zh-en and en-mr, while achieving results very close to the best baselines on other test sets. It even slightly surpasses the explicit reasoning performance on zh-en span-level F1 and en-mr word-level MCC. These results suggest that our training framework not only teaches the model to reason explicitly but also cultivates an internal intuition, enabling computationally efficient option for fine-grained prediction.

\paragraph{RIEQE activates the inherent linguistic capabilities of LLMs.}
Notably, when handling the low-resource language Marathi (mr), the untrained LLM exhibits severe limitations, with extremely low word-level MCC scores. However, after training with RIEQE, the MCC score increases dramatically, reaching as high as 0.2868 (RIEQE-NonThinking). This result surpasses \textit{DCSQE} and \textit{CometKiwi-23}, both of which were specifically trained on large-scale en-mr parallel corpora and QE datasets. Such a substantial performance gap strongly demonstrates that our method effectively activates the multilingual understanding capabilities within LLMs and transfers them to downstream fine-grained evaluation tasks.

\section{Collaboration between Implicit and Explicit Reasoning}

RIEQE enables an explicit reasoning model to simultaneously acquire implicit reasoning capabilities. In this section, we explain how task decomposition and \textit{NonThinking-SFT} facilitate implicit reasoning, and how explicit reasoning both benefits from and provides feedback to it.

\subsection{SubTask v.s. WholeTask Performance}
\label{sec:wholetask}

In this section, we empirically demonstrate that task decomposition yields substantial performance advantages over applying \textit{NonThinking-SFT} directly to the full fine-grained QE task.

Task decomposition enables reliable implicit reasoning by breaking the original task into sufficiently straightforward subtasks, where direct LLM responses become trustworthy. In contrast, the full task requires the model to simultaneously perform error detection over dozens of words and severity judgment within a single query. Under such settings, implicit reasoning within a limited number of model layers is unlikely to suffice; the model may instead rely on superficial patterns and shortcuts \citep{lin-etal-2025-implicit}. 
% Moreover, the available data scale and training budget fall far short of the regime required for emergent implicit multi-hop reasoning through grokking \citep{wang2024grokking}. 
We refer to this non-decomposed setting as \textbf{WholeTask}.

Of the 47k en-de training pairs, we used 1k for RLVR and the remaining 46k for SFT. The \textbf{SubTask} setting follows the same decomposition strategy as \S\ref{sec:taskDecomposition}. In the \textbf{WholeTask} setting, since task decomposition is removed, it contains fewer training instances. We therefore consider two variants: one with the same number of training epochs as the SubTask setting, and another with the same total number of training samples. The training setup remains the same as in \S\ref{sec:experiments}.

As shown in Table \ref{tab:subtask-vs-wholetask}, task decomposition brings significant performance gains, demonstrating that simpler and more direct subtasks are considerably more suitable for model learning.

\begin{table}[htbp]
\centering
\footnotesize
\begin{tabular}{lcc}
\hline
\multirow{2}{*}{Trainset} & \multicolumn{2}{c}{Word-level} \\
\cmidrule(lr){2-3} 
 & MCC & F1 \\
\hline
WholeTask (A)   & 0.2101 & 0.2623 \\
WholeTask (B) & 0.2320 & 0.2869 \\
SubTask     & \textbf{0.2674} & \textbf{0.3160} \\
\hline
\end{tabular}
\caption{Comparison between WholeTask and SubTask \textit{NonThinking} training and tested on en-de-2023. "(A)" means the same training epochs as SubTask, and "(B)" means the same amount of training samples.}
\label{tab:subtask-vs-wholetask}
\end{table}

\subsection{\textit{NonThinking} v.s. \textit{Thinking SFT} Performance}
\label{sec:nonthinkvsthink}
The prevailing paradigm is to first conduct SFT on synthetic long-CoT data~\citep{deepseekai2025deepseekv32pushingfrontieropen,ren2026rethinkinggeneralizationreasoningsft}, followed by reinforcement learning. We propose that for QE, a task that does not require particularly complex reasoning chains especially after task decomposition, \textit{NonThinking-SFT} can better enhance the model’s implicit reasoning tendency and, when combined with RLVR, lead to stronger performance.

To verify this, we design a controlled comparison study. We still use the en-de training set and still use 1k QE pairs for RLVR. The SFT data here contains only 10k samples, considering the substantial computational cost to synthesize reasoning chains. We use Qwen3-235B-A22B~\citep{yang2025qwen3technicalreport} in Thinking mode to generate ten responses for each training sample and retain responses according to 2 different selection strategies. In the first setting, only one correct response for each prompt is kept, while samples with all ten generations incorrect are discarded. After filtering, approximately half of the samples are retained. This setting is denoted as \textbf{Thinking-correct}. In the second setting, for each instance, we select the response that is closest to the correct answer among all sampled outputs. For example, if all the predicted error spans do not exactly match the ground truth, the response with the highest overlap is retained. This allows all prompts to be preserved, and we denote this setting as \textbf{Thinking-all}. Correspondingly, by removing the chain-of-thought content from these two SFT training sets, we construct two additional datasets, namely \textbf{NonThinking-correct} and \textbf{NonThinking-all}.

\begin{table}[htbp]
\centering
\footnotesize
% \small
\begin{tabular}{llcc}
\hline
\multirow{2}{*}{Testset} & \multirow{2}{*}{Trainset} & \multicolumn{2}{c}{Word-level} \\
 &  & MCC & F1 \\
\hline
\multirow{9}{*}{en-de}
& Qwen3-4B-Thinking-2507 & 0.1733 & 0.2226 \\
& \graycell{+ RLVR 1k} & \graycell{0.1882} & \graycell{0.1890} \\
% \cdashline{2-4}
& NonThinking-all      & 0.1953 & 0.2316 \\
& \graycell{+ RLVR 1k}         & \graycell{\textbf{0.2293}} & \graycell{0.2755} \\
& Thinking-all         & 0.1918 & 0.2439 \\
& \graycell{+ RLVR 1k}         & \graycell{\underline{0.2147}} & \graycell{0.2382} \\
& NonThinking-correct  & 0.1771 & 0.1948 \\
& \graycell{+ RLVR 1k}         & \graycell{0.2118} & \graycell{0.2420} \\
& Thinking-correct     & 0.1968 & 0.2329 \\
& \graycell{+ RLVR 1k}         & \graycell{0.2085} & \graycell{0.2375} \\
\hline
\end{tabular}
\caption{NonThinking-SFT v.s. Thinking-SFT performance on en-de testset at word-level. Rows with gray background represent model performance after applying the same RLVR procedure, using the preceding row as the base model.}
\label{tab:noncotvscot}
\end{table}

We perform SFT on each of the four datasets, followed by the same RLVR procedure. The results are shown in Table \ref{tab:noncotvscot}. Compared to RLVR-only (the first row), all SFT settings improve the scores. Although \textbf{NonThinking-all} does not achieve the best results after SFT alone, it delivers the strongest performance after RLVR across all four settings. Even \textbf{NonThinking-correct} shows consistent advantages over \textbf{Thinking-correct}. Since Qwen3-4B is distilled from Qwen3-235B-A22B, distribution mismatch can be ruled out as a confounding factor~\citep{yang2025qwen3technicalreport}. One possible explanation is that explicit reasoning trajectories inevitably contain  suboptimal or even wrong intermediate steps, which can influence the student model's performance. The implications for implicit reasoning ability are further analyzed in the next subsection. 

These results indicate that our method is not merely an efficient compromise for training and synthetic data generation bottlenecks, but can also achieve superior performance.

\subsection{Evidence of Enhanced Implicit Reasoning Tendency}

\begin{figure}
    \centering
    \includegraphics[width=\linewidth]{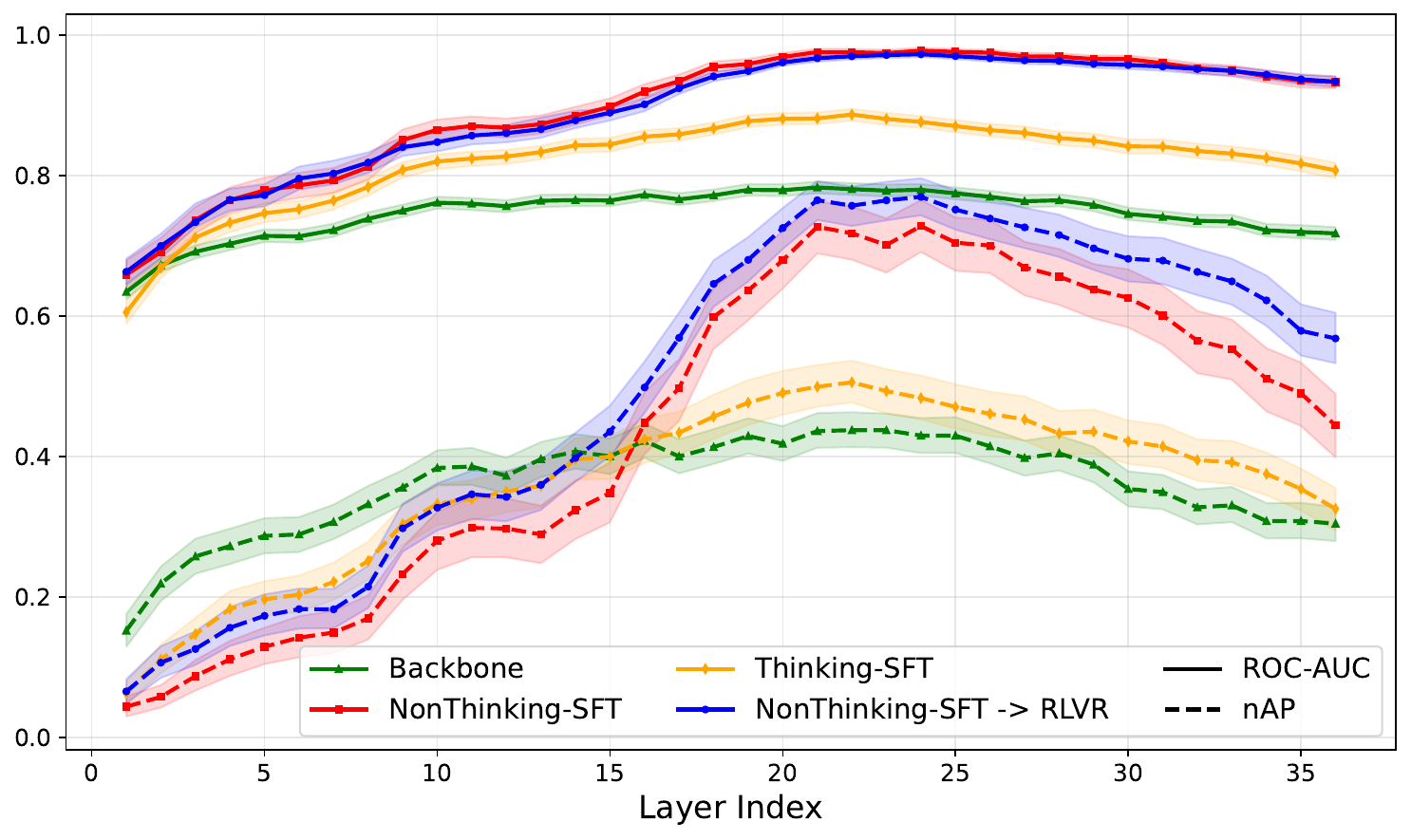}
    \caption{The discriminability scores of the model’s hidden states for teacher-forced translation sentences under four different settings. The random baselines for ROC-AUC and nAP are 0.5 and 0. ``Backbone'' denotes the untrained Qwen3-4B-Thinking-2507 model.}
    \label{fig:discriminability}
\end{figure}

Beyond avoiding the noise introduced by distillation, we further investigate how \textit{NonThinking-SFT} enhances implicit reasoning tendency. Specifically, we examine whether the model implicitly forms judgments when processing translation pairs in the input. If so, this suggests that the model plans ahead for subsequent output while reading the question~\citep{wu2024do,lindsey2025biology}, which may also benefit subsequent explicit reasoning.

% We train MLP probes that take the hidden states of individual translation tokens at each transformer layer as input and predict whether a token will later be classified as an error.
We train MLP probes to predict whether a translation token will later be classified as an error in the actual output.
Higher prediction accuracy indicates that the model has already encoded error-related information into the hidden states while reading the input question. The metrics on this imbalanced binary classification task are ROC-AUC and normalized Average Precision (nAP). ROC-AUC measures the model's overall ranking ability and is insensitive to class imbalance. Average Precision (AP) is defined as the area under the Precision–Recall curve; nAP normalizes AP by the random baseline, enabling fair comparisons. In particular, nAP is more informative under extreme class imbalance (such as in QE, where error tokens are sparse) since it focuses on precision–recall trade-offs for the positive class. We experiment on models trained with \textit{Thinking-SFT}, \textit{NonThinking-SFT} and \textit{NonThinking-SFT$\rightarrow$RLVR}. The other MLP training and evaluating details are in Appendix \ref{app:discriminability}.

\textbf{\textit{NonThinking-SFT} makes the model’s hidden states more separable.} As shown in Figure \ref{fig:discriminability}, the original backbone model has significantly lower ROC-AUC than the SFT model. The increased separability of hidden states suggests that the model is more inclined to encode error-related information in its latent space, reflecting stronger implicit reasoning tendency.

\textbf{\textit{Thinking-RLVR} also enhances implicit reasoning.} Although the ROC-AUC curves before and after \textit{Thinking-RLVR} are very close, the nAP curves reveal that further \textit{Thinking-RLVR} makes the hidden states more separable across all layers. This indicates that implicit reasoning is further strengthened by RLVR with reasoning explorations. 
% Although the nonCoT-RL model exhibits higher separability than the standard RLVR model in some layers, the results in Table \ref{tab:nonCOTRL} show that its overall performance is not optimal due to the lack of explicit CoT training.

% \begin{table}[ht]
%     \centering
%     \begin{tabular}{lcc}
%         \toprule
%          & Think & nonThink \\
%         \midrule
%         RLVR & 0.2858 & 0.2682 \\
%         nonCOT-RLVR & 0.2545 & 0.2408 \\
%         \bottomrule
%     \end{tabular}
%     \caption{Performance comparison w/ and w/o COT during RLVR. "LP" stands for "Language Pair".}
%     \label{tab:nonCOTRL}
% \end{table}

\subsection{Consistency Between Implicit Predictions and Explicit Reasoning Paths}

We analyze the consistency between implicit and explicit reasoning for the SFT and RLVR models under \textit{NonThinking-all} and \textit{Thinking-all} from Section~\ref{sec:nonthinkvsthink}. We examine both the consistency between the explicit reasoning result (Final Span) and the implicit reasoning result, and the consistency between the spans attended to during the explicit reasoning chain (Focus Span) and the implicit result. We use GPT-5.5 to identify which spans are attended to and discussed in each reasoning chain.

\begin{figure}
    \centering
    \includegraphics[width=\linewidth]{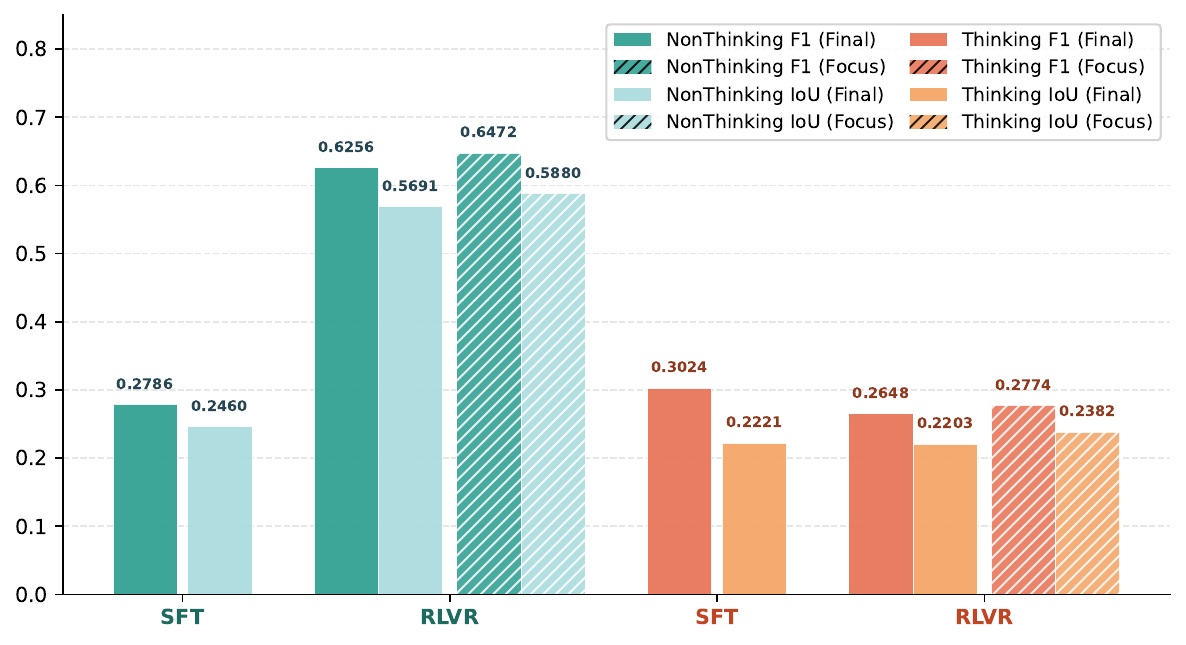}
    \caption{Consistency between explicit and implicit reasoning. ``NonThinking'' and ``Thinking'' refer to models that underwent \textit{NonThinking-SFT} and \textit{Thinking-SFT}, respectively.}
    \label{fig:consistencyimpexp}
\end{figure}

The results are shown in Figure~\ref{fig:consistencyimpexp}. We observe that although the model after SFT exhibits a much lower consistency rate between its explicit and implicit predictions, the \textit{NonThinking-SFT} model after RLVR exhibits much higher consistency.
% up to 0.6472 IoU. 
This suggests that during \textit{Thinking-RLVR}, \textbf{the model is guided by its acquired implicit reasoning capability}, which is a significant difference from \textit{Thinking-SFT}, leading to stronger performance.

\section{Related Work}

Encoder-based models have consistently achieved the best performance in QE, as they excel at token-level fine-grained discrimination and regression fitting of quality scores. A representative lineage of such approaches began with a reference-based method COMET \citep{rei-etal-2020-comet}, and then evolved into reference-free model CometKiwi \citep{rei-etal-2022-cometkiwi,rei-etal-2023-scaling} with scaled training data. xCOMET \citep{guerreiro-etal-2024-xcomet} seamlessly integrates reference-based and reference-free evaluation while highlighting span-level error detection. Another line of work focuses on high quality data synthesis for QE, such as MQMQE \citep{geng-etal-2023-unify} and DCSQE \citep{geng-etal-2025-alleviating}.

Some studies have attempted to apply LLMs to QE tasks, including InstructScore~\citep{xu-etal-2023-instructscore} and GEMBA-MQM~\citep{kocmi-federmann-2023-gemba}. However, their performance still lag behind encoder-based models. Results from WMT25~\citep{wmt2025qefindings} show that advanced LRM-based methods have achieved advantages in sentence-level scoring with superior reasoning ability, such as TASER~\citep{maheswaran-etal-2025-taser} and GEMBA-v2~\citep{junczys-dowmunt-2025-gemba}. Nevertheless, for fine-grained QE tasks, none of the submitted systems surpass the baseline model xCOMET, even when leveraging powerful models such as OpenAI-o3/o4-mini~\citep{o3systemcard}. %This indicates that fine-grained QE remains highly challenging for LLMs and LRMs.

\section{Conclusion}

In this paper, we propose RIEQE, a simple two-stage training framework that unlocks fine-grained QE in LRMs through mutually boosting of implicit and explicit reasoning. By decomposing the complex QE task into simpler subtasks and applying \textit{NonThink-SFT} followed by \textit{Thinking-RLVR}, our method achieves state-of-the-art explicit reasoning performance and competitive implicit reasoning capability. Further analyses reveal that NonThinking-SFT forces the model to compress reasoning capabilities into a layer-wise form. This strong implicit reasoning capability provides a crucial foundation for subsequent explicit reasoning training, while explicit reasoning optimization in turn further enhances implicit reasoning tendency. These findings suggest that the implicit and explicit reasoning capabilities of LRMs are deeply intertwined and can mutually reinforce each other through appropriate training, offering a new research perspective on integrating implicit and explicit thinking in modern LRMs.

% \clearpage
\section*{Limitations}
This work has several limitations. 
First, the study does not systematically investigate the optimal data composition or the impact of training hyperparameters, which may leave the presented results suboptimal.
Second, the experimental analysis of this implicit and explicit co-evolution is relatively limited, with most results focusing on final task performance rather than directly probing how and why the two reasoning modes interact during training. 
Third, the RIEQE framework is designed for fine-grained QE task, its applicability to other reasoning tasks remains unexplored. 

\section*{Acknowledgement}
We would like to thank the anonymous reviewers for their insightful comments. Shujian Huang is the corresponding author. This work is supported by National Science Foundation of China (No. 62376116), the Fundamental Research Funds for the Central Universities (No. 2024300507), Fundamental and Interdisciplinary Disciplines Breakthrough Plan of the Ministry of Education of China (No. JYB2025XDXM118).

\bibliography{custom}

@inproceedings{geng-etal-2025-alleviating,
    title = "Alleviating Distribution Shift in Synthetic Data for Machine Translation Quality Estimation",
    author = "Geng, Xiang  and
      Lai, Zhejian  and
      Chen, Jiajun  and
      Yang, Hao  and
      Huang, Shujian",
    editor = "Che, Wanxiang  and
      Nabende, Joyce  and
      Shutova, Ekaterina  and
      Pilehvar, Mohammad Taher",
    booktitle = "Proceedings of the 63rd Annual Meeting of the Association for Computational Linguistics (Volume 1: Long Papers)",
    month = jul,
    year = "2025",
    address = "Vienna, Austria",
    publisher = "Association for Computational Linguistics",
    url = "https://aclanthology.org/2025.acl-long.373/",
    doi = "10.18653/v1/2025.acl-long.373",
    pages = "7546--7560",
    ISBN = "979-8-89176-251-0"
}

@inproceedings{wmt2025qefindings,
    title = "Findings of the {WMT}25 Shared Task on Automated Translation Evaluation Systems: Linguistic Diversity is Challenging and References Still Help",
    author = "Lavie, Alon  and
      Hanneman, Greg  and
      Agrawal, Sweta  and
      Kanojia, Diptesh  and
      Lo, Chi-Kiu  and
      Zouhar, Vil{\'e}m  and
      Blain, Frederic  and
      Zerva, Chrysoula  and
      Avramidis, Eleftherios  and
      Deoghare, Sourabh  and
      Sindhujan, Archchana  and
      Wang, Jiayi  and
      Adelani, David Ifeoluwa  and
      Thompson, Brian  and
      Kocmi, Tom  and
      Freitag, Markus  and
      Deutsch, Daniel",
    editor = "Haddow, Barry  and
      Kocmi, Tom  and
      Koehn, Philipp  and
      Monz, Christof",
    booktitle = "Proceedings of the Tenth Conference on Machine Translation",
    month = nov,
    year = "2025",
    address = "Suzhou, China",
    publisher = "Association for Computational Linguistics",
    url = "https://aclanthology.org/2025.wmt-1.24/",
    doi = "10.18653/v1/2025.wmt-1.24",
    pages = "436--483",
    ISBN = "979-8-89176-341-8"
}

@inproceedings{wmt2025mtfindings,
    title = "Findings of the {WMT}25 General Machine Translation Shared Task: Time to Stop Evaluating on Easy Test Sets",
    author = "Kocmi, Tom  and
      Artemova, Ekaterina  and
      Avramidis, Eleftherios  and
      Bawden, Rachel  and
      Bojar, Ond{\v{r}}ej  and
      Dranch, Konstantin  and
      Dvorkovich, Anton  and
      Dukanov, Sergey  and
      Fishel, Mark  and
      Freitag, Markus  and
      Gowda, Thamme  and
      Grundkiewicz, Roman  and
      Haddow, Barry  and
      Karpinska, Marzena  and
      Koehn, Philipp  and
      Lakougna, Howard  and
      Lundin, Jessica  and
      Monz, Christof  and
      Murray, Kenton  and
      Nagata, Masaaki  and
      Perrella, Stefano  and
      Proietti, Lorenzo  and
      Popel, Martin  and
      Popovi{\'c}, Maja  and
      Riley, Parker  and
      Shmatova, Mariya  and
      Steingr{\'i}msson, Steinth{\'o}r  and
      Yankovskaya, Lisa  and
      Zouhar, Vil{\'e}m",
    editor = "Haddow, Barry  and
      Kocmi, Tom  and
      Koehn, Philipp  and
      Monz, Christof",
    booktitle = "Proceedings of the Tenth Conference on Machine Translation",
    month = nov,
    year = "2025",
    address = "Suzhou, China",
    publisher = "Association for Computational Linguistics",
    url = "https://aclanthology.org/2025.wmt-1.22/",
    doi = "10.18653/v1/2025.wmt-1.22",
    pages = "355--413",
    ISBN = "979-8-89176-341-8"
}

@inproceedings{wmt2024qefindings,
    title = "Findings of the Quality Estimation Shared Task at {WMT} 2024: Are {LLM}s Closing the Gap in {QE}?",
    author = "Zerva, Chrysoula  and
      Blain, Frederic  and
      C. De Souza, Jos{\'e} G.  and
      Kanojia, Diptesh  and
      Deoghare, Sourabh  and
      Guerreiro, Nuno M.  and
      Attanasio, Giuseppe  and
      Rei, Ricardo  and
      Orasan, Constantin  and
      Negri, Matteo  and
      Turchi, Marco  and
      Chatterjee, Rajen  and
      Bhattacharyya, Pushpak  and
      Freitag, Markus  and
      Martins, Andr{\'e}",
    editor = "Haddow, Barry  and
      Kocmi, Tom  and
      Koehn, Philipp  and
      Monz, Christof",
    booktitle = "Proceedings of the Ninth Conference on Machine Translation",
    month = nov,
    year = "2024",
    address = "Miami, Florida, USA",
    publisher = "Association for Computational Linguistics",
    url = "https://aclanthology.org/2024.wmt-1.3/",
    doi = "10.18653/v1/2024.wmt-1.3",
    pages = "82--109"
}

@inproceedings{huang-etal-2024-lost,
    title = "Lost in the Source Language: How Large Language Models Evaluate the Quality of Machine Translation",
    author = "Huang, Xu  and
      Zhang, Zhirui  and
      Geng, Xiang  and
      Du, Yichao  and
      Chen, Jiajun  and
      Huang, Shujian",
    editor = "Ku, Lun-Wei  and
      Martins, Andre  and
      Srikumar, Vivek",
    booktitle = "Findings of the Association for Computational Linguistics: ACL 2024",
    month = aug,
    year = "2024",
    address = "Bangkok, Thailand",
    publisher = "Association for Computational Linguistics",
    url = "https://aclanthology.org/2024.findings-acl.211/",
    doi = "10.18653/v1/2024.findings-acl.211",
    pages = "3546--3562"
}

@inproceedings{
dang2026the,
title={The First Impression Problem: Internal Bias Triggers Overthinking in Reasoning Models},
author={Renfei Dang and Zhening Li and Shujian Huang and Jiajun Chen},
booktitle={The Fourteenth International Conference on Learning Representations},
year={2026},
url={https://openreview.net/forum?id=2PP70tFY0S}
}

@misc{chen2025reasoningmodelsdontsay,
      title={Reasoning Models Don't Always Say What They Think}, 
      author={Yanda Chen and Joe Benton and Ansh Radhakrishnan and Jonathan Uesato and Carson Denison and John Schulman and Arushi Somani and Peter Hase and Misha Wagner and Fabien Roger and Vlad Mikulik and Samuel R. Bowman and Jan Leike and Jared Kaplan and Ethan Perez},
      year={2025},
      eprint={2505.05410},
      archivePrefix={arXiv},
      primaryClass={cs.CL},
      url={https://arxiv.org/abs/2505.05410}, 
}

@inproceedings{
arcuschin2025chainofthought,
title={Chain-of-Thought Reasoning in the Wild is not Always Faithful},
author={Iv{\'a}n Arcuschin and Jett Janiak and Robert Krzyzanowski and Senthooran Rajamanoharan and Neel Nanda and Arthur Conmy},
booktitle={Workshop on Reasoning and Planning for Large Language Models},
year={2025},
url={https://openreview.net/forum?id=L8094Whth0}
}

@inproceedings{tyen-etal-2024-llms,
    title = "{LLM}s cannot find reasoning errors, but can correct them given the error location",
    author = "Tyen, Gladys  and
      Mansoor, Hassan  and
      Carbune, Victor  and
      Chen, Peter  and
      Mak, Tony",
    editor = "Ku, Lun-Wei  and
      Martins, Andre  and
      Srikumar, Vivek",
    booktitle = "Findings of the Association for Computational Linguistics: ACL 2024",
    month = aug,
    year = "2024",
    address = "Bangkok, Thailand",
    publisher = "Association for Computational Linguistics",
    url = "https://aclanthology.org/2024.findings-acl.826/",
    doi = "10.18653/v1/2024.findings-acl.826",
    pages = "13894--13908"
}

@inproceedings{rei-etal-2020-comet,
    title = "{COMET}: A Neural Framework for {MT} Evaluation",
    author = "Rei, Ricardo  and
      Stewart, Craig  and
      Farinha, Ana C  and
      Lavie, Alon",
    editor = "Webber, Bonnie  and
      Cohn, Trevor  and
      He, Yulan  and
      Liu, Yang",
    booktitle = "Proceedings of the 2020 Conference on Empirical Methods in Natural Language Processing (EMNLP)",
    month = nov,
    year = "2020",
    address = "Online",
    publisher = "Association for Computational Linguistics",
    url = "https://aclanthology.org/2020.emnlp-main.213/",
    doi = "10.18653/v1/2020.emnlp-main.213",
    pages = "2685--2702"
}

@inproceedings{rei-etal-2022-cometkiwi,
    title = "{C}omet{K}iwi: {IST}-Unbabel 2022 Submission for the Quality Estimation Shared Task",
    author = "Rei, Ricardo  and
      Treviso, Marcos  and
      Guerreiro, Nuno M.  and
      Zerva, Chrysoula  and
      Farinha, Ana C  and
      Maroti, Christine  and
      C. de Souza, Jos{\'e} G.  and
      Glushkova, Taisiya  and
      Alves, Duarte  and
      Coheur, Luisa  and
      Lavie, Alon  and
      Martins, Andr{\'e} F. T.",
    editor = {Koehn, Philipp  and
      Barrault, Lo{\"i}c  and
      Bojar, Ond{\v{r}}ej  and
      Bougares, Fethi  and
      Chatterjee, Rajen  and
      Costa-juss{\`a}, Marta R.  and
      Federmann, Christian  and
      Fishel, Mark  and
      Fraser, Alexander  and
      Freitag, Markus  and
      Graham, Yvette  and
      Grundkiewicz, Roman  and
      Guzman, Paco  and
      Haddow, Barry  and
      Huck, Matthias  and
      Jimeno Yepes, Antonio  and
      Kocmi, Tom  and
      Martins, Andr{\'e}  and
      Morishita, Makoto  and
      Monz, Christof  and
      Nagata, Masaaki  and
      Nakazawa, Toshiaki  and
      Negri, Matteo  and
      N{\'e}v{\'e}ol, Aur{\'e}lie  and
      Neves, Mariana  and
      Popel, Martin  and
      Turchi, Marco  and
      Zampieri, Marcos},
    booktitle = "Proceedings of the Seventh Conference on Machine Translation (WMT)",
    month = dec,
    year = "2022",
    address = "Abu Dhabi, United Arab Emirates (Hybrid)",
    publisher = "Association for Computational Linguistics",
    url = "https://aclanthology.org/2022.wmt-1.60/",
    doi = "10.18653/v1/2022.wmt-1.60",
    pages = "634--645"
}

@inproceedings{rei-etal-2023-scaling,
    title = "Scaling up {C}omet{K}iwi: Unbabel-{IST} 2023 Submission for the Quality Estimation Shared Task",
    author = "Rei, Ricardo  and
      Guerreiro, Nuno M.  and
      Pombal, Jos{\'e}  and
      van Stigt, Daan  and
      Treviso, Marcos  and
      Coheur, Luisa  and
      C. de Souza, Jos{\'e} G.  and
      Martins, Andr{\'e} F. T.",
    editor = "Koehn, Philipp  and
      Haddow, Barry  and
      Kocmi, Tom  and
      Monz, Christof",
    booktitle = "Proceedings of the Eighth Conference on Machine Translation",
    month = dec,
    year = "2023",
    address = "Singapore",
    publisher = "Association for Computational Linguistics",
    url = "https://aclanthology.org/2023.wmt-1.73/",
    doi = "10.18653/v1/2023.wmt-1.73",
    pages = "841--848"
}

@article{guerreiro-etal-2024-xcomet,
    title = "x{COMET}: Transparent Machine Translation Evaluation through Fine-grained Error Detection",
    author = "Guerreiro, Nuno M.  and
      Rei, Ricardo  and
      Stigt, Daan van  and
      Coheur, Luisa  and
      Colombo, Pierre  and
      Martins, Andr{\'e} F. T.",
    journal = "Transactions of the Association for Computational Linguistics",
    volume = "12",
    year = "2024",
    address = "Cambridge, MA",
    publisher = "MIT Press",
    url = "https://aclanthology.org/2024.tacl-1.54/",
    doi = "10.1162/tacl_a_00683",
    pages = "979--995"
}

@inproceedings{geng-etal-2023-unify,
    title = "Unify Word-level and Span-level Tasks: {NJUNLP}{'}s Participation for the {WMT}2023 Quality Estimation Shared Task",
    author = "Geng, Xiang  and
      Lai, Zhejian  and
      Zhang, Yu  and
      Tao, Shimin  and
      Yang, Hao  and
      Chen, Jiajun  and
      Huang, Shujian",
    editor = "Koehn, Philipp  and
      Haddow, Barry  and
      Kocmi, Tom  and
      Monz, Christof",
    booktitle = "Proceedings of the Eighth Conference on Machine Translation",
    month = dec,
    year = "2023",
    address = "Singapore",
    publisher = "Association for Computational Linguistics",
    url = "https://aclanthology.org/2023.wmt-1.71/",
    doi = "10.18653/v1/2023.wmt-1.71",
    pages = "829--834"
}

@inproceedings{xu-etal-2023-instructscore,
    title = "{INSTRUCTSCORE}: Towards Explainable Text Generation Evaluation with Automatic Feedback",
    author = "Xu, Wenda  and
      Wang, Danqing  and
      Pan, Liangming  and
      Song, Zhenqiao  and
      Freitag, Markus  and
      Wang, William  and
      Li, Lei",
    editor = "Bouamor, Houda  and
      Pino, Juan  and
      Bali, Kalika",
    booktitle = "Proceedings of the 2023 Conference on Empirical Methods in Natural Language Processing",
    month = dec,
    year = "2023",
    address = "Singapore",
    publisher = "Association for Computational Linguistics",
    url = "https://aclanthology.org/2023.emnlp-main.365/",
    doi = "10.18653/v1/2023.emnlp-main.365",
    pages = "5967--5994"
}

@inproceedings{kocmi-federmann-2023-gemba,
    title = "{GEMBA}-{MQM}: Detecting Translation Quality Error Spans with {GPT}-4",
    author = "Kocmi, Tom  and
      Federmann, Christian",
    editor = "Koehn, Philipp  and
      Haddow, Barry  and
      Kocmi, Tom  and
      Monz, Christof",
    booktitle = "Proceedings of the Eighth Conference on Machine Translation",
    month = dec,
    year = "2023",
    address = "Singapore",
    publisher = "Association for Computational Linguistics",
    url = "https://aclanthology.org/2023.wmt-1.64/",
    doi = "10.18653/v1/2023.wmt-1.64",
    pages = "768--775"
}

@inproceedings{maheswaran-etal-2025-taser,
    title = "{TASER}: Translation Assessment via Systematic Evaluation and Reasoning",
    author = "Maheswaran, Monishwaran  and
      Carini, Marco  and
      Federmann, Christian  and
      Diaz, Tony",
    editor = "Haddow, Barry  and
      Kocmi, Tom  and
      Koehn, Philipp  and
      Monz, Christof",
    booktitle = "Proceedings of the Tenth Conference on Machine Translation",
    month = nov,
    year = "2025",
    address = "Suzhou, China",
    publisher = "Association for Computational Linguistics",
    url = "https://aclanthology.org/2025.wmt-1.76/",
    doi = "10.18653/v1/2025.wmt-1.76",
    pages = "1004--1010",
    ISBN = "979-8-89176-341-8"
}

@inproceedings{junczys-dowmunt-2025-gemba,
    title = "{GEMBA} V2: Ten Judgments Are Better Than One",
    author = "Junczys-Dowmunt, Marcin",
    editor = "Haddow, Barry  and
      Kocmi, Tom  and
      Koehn, Philipp  and
      Monz, Christof",
    booktitle = "Proceedings of the Tenth Conference on Machine Translation",
    month = nov,
    year = "2025",
    address = "Suzhou, China",
    publisher = "Association for Computational Linguistics",
    url = "https://aclanthology.org/2025.wmt-1.67/",
    doi = "10.18653/v1/2025.wmt-1.67",
    pages = "926--933",
    ISBN = "979-8-89176-341-8"
}

@inproceedings{chen-etal-2025-beyond,
    title = "Beyond the Surface: Measuring Self-Preference in {LLM} Judgments",
    author = "Chen, Zhi-Yuan  and
      Wang, Hao  and
      Zhang, Xinyu  and
      Hu, Enrui  and
      Lin, Yankai",
    editor = "Christodoulopoulos, Christos  and
      Chakraborty, Tanmoy  and
      Rose, Carolyn  and
      Peng, Violet",
    booktitle = "Proceedings of the 2025 Conference on Empirical Methods in Natural Language Processing",
    month = nov,
    year = "2025",
    address = "Suzhou, China",
    publisher = "Association for Computational Linguistics",
    url = "https://aclanthology.org/2025.emnlp-main.86/",
    doi = "10.18653/v1/2025.emnlp-main.86",
    pages = "1653--1672",
    ISBN = "979-8-89176-332-6"
}

@misc{zhang2026penalizinglengthuncoveringsystematic,
      title={Penalizing Length: Uncovering Systematic Bias in Quality Estimation Metrics}, 
      author={Yilin Zhang and Wenda Xu and Zhongtao Liu and Tetsuji Nakagawa and Markus Freitag},
      year={2026},
      eprint={2510.22028},
      archivePrefix={arXiv},
      primaryClass={cs.CL},
      url={https://arxiv.org/abs/2510.22028}, 
}

@inproceedings{wang-etal-2024-large-language-models-fair,
    title = "Large Language Models are not Fair Evaluators",
    author = "Wang, Peiyi  and
      Li, Lei  and
      Chen, Liang  and
      Cai, Zefan  and
      Zhu, Dawei  and
      Lin, Binghuai  and
      Cao, Yunbo  and
      Kong, Lingpeng  and
      Liu, Qi  and
      Liu, Tianyu  and
      Sui, Zhifang",
    editor = "Ku, Lun-Wei  and
      Martins, Andre  and
      Srikumar, Vivek",
    booktitle = "Proceedings of the 62nd Annual Meeting of the Association for Computational Linguistics (Volume 1: Long Papers)",
    month = aug,
    year = "2024",
    address = "Bangkok, Thailand",
    publisher = "Association for Computational Linguistics",
    url = "https://aclanthology.org/2024.acl-long.511/",
    doi = "10.18653/v1/2024.acl-long.511",
    pages = "9440--9450"
}

@misc{zhou2026fairnessfluencyinvestigationlanguage,
      title={Fairness or Fluency? An Investigation into Language Bias of Pairwise LLM-as-a-Judge}, 
      author={Xiaolin Zhou and Zheng Luo and Yicheng Gao and Qixuan Chen and Xiyang Hu and Yue Zhao and Ruishan Liu},
      year={2026},
      eprint={2601.13649},
      archivePrefix={arXiv},
      primaryClass={cs.CL},
      url={https://arxiv.org/abs/2601.13649}, 
}

@inproceedings{blain-etal-2023-findings,
    title = "Findings of the {WMT} 2023 Shared Task on Quality Estimation",
    author = "Blain, Frederic  and
      Zerva, Chrysoula  and
      Rei, Ricardo  and
      Guerreiro, Nuno M.  and
      Kanojia, Diptesh  and
      C. de Souza, Jos{\'e} G.  and
      Silva, Beatriz  and
      Vaz, T{\^a}nia  and
      Jingxuan, Yan  and
      Azadi, Fatemeh  and
      Orasan, Constantin  and
      Martins, Andr{\'e}",
    editor = "Koehn, Philipp  and
      Haddow, Barry  and
      Kocmi, Tom  and
      Monz, Christof",
    booktitle = "Proceedings of the Eighth Conference on Machine Translation",
    month = dec,
    year = "2023",
    address = "Singapore",
    publisher = "Association for Computational Linguistics",
    url = "https://aclanthology.org/2023.wmt-1.52/",
    doi = "10.18653/v1/2023.wmt-1.52",
    pages = "629--653"
}

@inproceedings{conneau-etal-2020-unsupervised,
    title = "Unsupervised Cross-lingual Representation Learning at Scale",
    author = "Conneau, Alexis  and
      Khandelwal, Kartikay  and
      Goyal, Naman  and
      Chaudhary, Vishrav  and
      Wenzek, Guillaume  and
      Guzm{\'a}n, Francisco  and
      Grave, Edouard  and
      Ott, Myle  and
      Zettlemoyer, Luke  and
      Stoyanov, Veselin",
    editor = "Jurafsky, Dan  and
      Chai, Joyce  and
      Schluter, Natalie  and
      Tetreault, Joel",
    booktitle = "Proceedings of the 58th Annual Meeting of the Association for Computational Linguistics",
    month = jul,
    year = "2020",
    address = "Online",
    publisher = "Association for Computational Linguistics",
    url = "https://aclanthology.org/2020.acl-main.747/",
    doi = "10.18653/v1/2020.acl-main.747",
    pages = "8440--8451"
}

@misc{deepseekai2025deepseekv32pushingfrontieropen,
      title={DeepSeek-V3.2: Pushing the Frontier of Open Large Language Models}, 
      author={DeepSeek-AI and Aixin Liu and Aoxue Mei and Bangcai Lin and Bing Xue and Bingxuan Wang and Bingzheng Xu and Bochao Wu and Bowei Zhang and Chaofan Lin and Chen Dong and Chengda Lu and Chenggang Zhao and Chengqi Deng and Chenhao Xu and Chong Ruan and Damai Dai and Daya Guo and Dejian Yang and Deli Chen and Erhang Li and Fangqi Zhou and Fangyun Lin and Fucong Dai and Guangbo Hao and Guanting Chen and Guowei Li and H. Zhang and Hanwei Xu and Hao Li and Haofen Liang and Haoran Wei and Haowei Zhang and Haowen Luo and Haozhe Ji and Honghui Ding and Hongxuan Tang and Huanqi Cao and Huazuo Gao and Hui Qu and Hui Zeng and Jialiang Huang and Jiashi Li and Jiaxin Xu and Jiewen Hu and Jingchang Chen and Jingting Xiang and Jingyang Yuan and Jingyuan Cheng and Jinhua Zhu and Jun Ran and Junguang Jiang and Junjie Qiu and Junlong Li and Junxiao Song and Kai Dong and Kaige Gao and Kang Guan and Kexin Huang and Kexing Zhou and Kezhao Huang and Kuai Yu and Lean Wang and Lecong Zhang and Lei Wang and Liang Zhao and Liangsheng Yin and Lihua Guo and Lingxiao Luo and Linwang Ma and Litong Wang and Liyue Zhang and M. S. Di and M. Y Xu and Mingchuan Zhang and Minghua Zhang and Minghui Tang and Mingxu Zhou and Panpan Huang and Peixin Cong and Peiyi Wang and Qiancheng Wang and Qihao Zhu and Qingyang Li and Qinyu Chen and Qiushi Du and Ruiling Xu and Ruiqi Ge and Ruisong Zhang and Ruizhe Pan and Runji Wang and Runqiu Yin and Runxin Xu and Ruomeng Shen and Ruoyu Zhang and S. H. Liu and Shanghao Lu and Shangyan Zhou and Shanhuang Chen and Shaofei Cai and Shaoyuan Chen and Shengding Hu and Shengyu Liu and Shiqiang Hu and Shirong Ma and Shiyu Wang and Shuiping Yu and Shunfeng Zhou and Shuting Pan and Songyang Zhou and Tao Ni and Tao Yun and Tian Pei and Tian Ye and Tianyuan Yue and Wangding Zeng and Wen Liu and Wenfeng Liang and Wenjie Pang and Wenjing Luo and Wenjun Gao and Wentao Zhang and Xi Gao and Xiangwen Wang and Xiao Bi and Xiaodong Liu and Xiaohan Wang and Xiaokang Chen and Xiaokang Zhang and Xiaotao Nie and Xin Cheng and Xin Liu and Xin Xie and Xingchao Liu and Xingkai Yu and Xingyou Li and Xinyu Yang and Xinyuan Li and Xu Chen and Xuecheng Su and Xuehai Pan and Xuheng Lin and Xuwei Fu and Y. Q. Wang and Yang Zhang and Yanhong Xu and Yanru Ma and Yao Li and Yao Li and Yao Zhao and Yaofeng Sun and Yaohui Wang and Yi Qian and Yi Yu and Yichao Zhang and Yifan Ding and Yifan Shi and Yiliang Xiong and Ying He and Ying Zhou and Yinmin Zhong and Yishi Piao and Yisong Wang and Yixiao Chen and Yixuan Tan and Yixuan Wei and Yiyang Ma and Yiyuan Liu and Yonglun Yang and Yongqiang Guo and Yongtong Wu and Yu Wu and Yuan Cheng and Yuan Ou and Yuanfan Xu and Yuduan Wang and Yue Gong and Yuhan Wu and Yuheng Zou and Yukun Li and Yunfan Xiong and Yuxiang Luo and Yuxiang You and Yuxuan Liu and Yuyang Zhou and Z. F. Wu and Z. Z. Ren and Zehua Zhao and Zehui Ren and Zhangli Sha and Zhe Fu and Zhean Xu and Zhenda Xie and Zhengyan Zhang and Zhewen Hao and Zhibin Gou and Zhicheng Ma and Zhigang Yan and Zhihong Shao and Zhixian Huang and Zhiyu Wu and Zhuoshu Li and Zhuping Zhang and Zian Xu and Zihao Wang and Zihui Gu and Zijia Zhu and Zilin Li and Zipeng Zhang and Ziwei Xie and Ziyi Gao and Zizheng Pan and Zongqing Yao and Bei Feng and Hui Li and J. L. Cai and Jiaqi Ni and Lei Xu and Meng Li and Ning Tian and R. J. Chen and R. L. Jin and S. S. Li and Shuang Zhou and Tianyu Sun and X. Q. Li and Xiangyue Jin and Xiaojin Shen and Xiaosha Chen and Xinnan Song and Xinyi Zhou and Y. X. Zhu and Yanping Huang and Yaohui Li and Yi Zheng and Yuchen Zhu and Yunxian Ma and Zhen Huang and Zhipeng Xu and Zhongyu Zhang and Dongjie Ji and Jian Liang and Jianzhong Guo and Jin Chen and Leyi Xia and Miaojun Wang and Mingming Li and Peng Zhang and Ruyi Chen and Shangmian Sun and Shaoqing Wu and Shengfeng Ye and T. Wang and W. L. Xiao and Wei An and Xianzu Wang and Xiaowen Sun and Xiaoxiang Wang and Ying Tang and Yukun Zha and Zekai Zhang and Zhe Ju and Zhen Zhang and Zihua Qu},
      year={2025},
      eprint={2512.02556},
      archivePrefix={arXiv},
      primaryClass={cs.CL},
      url={https://arxiv.org/abs/2512.02556}, 
}

@misc{ren2026rethinkinggeneralizationreasoningsft,
      title={Rethinking Generalization in Reasoning SFT: A Conditional Analysis on Optimization, Data, and Model Capability}, 
      author={Qihan Ren and Peng Wang and Ruikun Cai and Shuai Shao and Dadi Guo and Yuejin Xie and Yafu Li and Quanshi Zhang and Xia Hu and Jing Shao and Dongrui Liu},
      year={2026},
      eprint={2604.06628},
      archivePrefix={arXiv},
      primaryClass={cs.AI},
      url={https://arxiv.org/abs/2604.06628}, 
}

@misc{yang2025qwen3technicalreport,
      title={Qwen3 Technical Report}, 
      author={An Yang and Anfeng Li and Baosong Yang and Beichen Zhang and Binyuan Hui and Bo Zheng and Bowen Yu and Chang Gao and Chengen Huang and Chenxu Lv and Chujie Zheng and Dayiheng Liu and Fan Zhou and Fei Huang and Feng Hu and Hao Ge and Haoran Wei and Huan Lin and Jialong Tang and Jian Yang and Jianhong Tu and Jianwei Zhang and Jianxin Yang and Jiaxi Yang and Jing Zhou and Jingren Zhou and Junyang Lin and Kai Dang and Keqin Bao and Kexin Yang and Le Yu and Lianghao Deng and Mei Li and Mingfeng Xue and Mingze Li and Pei Zhang and Peng Wang and Qin Zhu and Rui Men and Ruize Gao and Shixuan Liu and Shuang Luo and Tianhao Li and Tianyi Tang and Wenbiao Yin and Xingzhang Ren and Xinyu Wang and Xinyu Zhang and Xuancheng Ren and Yang Fan and Yang Su and Yichang Zhang and Yinger Zhang and Yu Wan and Yuqiong Liu and Zekun Wang and Zeyu Cui and Zhenru Zhang and Zhipeng Zhou and Zihan Qiu},
      year={2025},
      eprint={2505.09388},
      archivePrefix={arXiv},
      primaryClass={cs.CL},
      url={https://arxiv.org/abs/2505.09388}, 
}

@inproceedings{lin-etal-2025-implicit,
    title = "Implicit Reasoning in Transformers is Reasoning through Shortcuts",
    author = "Lin, Tianhe  and
      Xie, Jian  and
      Yuan, Siyu  and
      Yang, Deqing",
    editor = "Che, Wanxiang  and
      Nabende, Joyce  and
      Shutova, Ekaterina  and
      Pilehvar, Mohammad Taher",
    booktitle = "Findings of the Association for Computational Linguistics: ACL 2025",
    month = jul,
    year = "2025",
    address = "Vienna, Austria",
    publisher = "Association for Computational Linguistics",
    url = "https://aclanthology.org/2025.findings-acl.493/",
    doi = "10.18653/v1/2025.findings-acl.493",
    pages = "9470--9487",
    ISBN = "979-8-89176-256-5"
}

@misc{shao2024deepseekmathpushinglimitsmathematical,
      title={DeepSeekMath: Pushing the Limits of Mathematical Reasoning in Open Language Models}, 
      author={Zhihong Shao and Peiyi Wang and Qihao Zhu and Runxin Xu and Junxiao Song and Xiao Bi and Haowei Zhang and Mingchuan Zhang and Y. K. Li and Y. Wu and Daya Guo},
      year={2024},
      eprint={2402.03300},
      archivePrefix={arXiv},
      primaryClass={cs.CL},
      url={https://arxiv.org/abs/2402.03300}, 
}

@inproceedings{
hu2022lora,
title={Lo{RA}: Low-Rank Adaptation of Large Language Models},
author={Edward J Hu and yelong shen and Phillip Wallis and Zeyuan Allen-Zhu and Yuanzhi Li and Shean Wang and Lu Wang and Weizhu Chen},
booktitle={International Conference on Learning Representations},
year={2022},
url={https://openreview.net/forum?id=nZeVKeeFYf9}
}

@inproceedings{zheng2024llamafactory,
  title={LlamaFactory: Unified Efficient Fine-Tuning of 100+ Language Models},
  author={Yaowei Zheng and Richong Zhang and Junhao Zhang and Yanhan Ye and Zheyan Luo and Zhangchi Feng and Yongqiang Ma},
  booktitle={Proceedings of the 62nd Annual Meeting of the Association for Computational Linguistics (Volume 3: System Demonstrations)},
  address={Bangkok, Thailand},
  publisher={Association for Computational Linguistics},
  year={2024},
  url={http://arxiv.org/abs/2403.13372}
}

@inproceedings{verl,
author = {Sheng, Guangming and Zhang, Chi and Ye, Zilingfeng and Wu, Xibin and Zhang, Wang and Zhang, Ru and Peng, Yanghua and Lin, Haibin and Wu, Chuan},
title = {HybridFlow: A Flexible and Efficient RLHF Framework},
year = {2025},
isbn = {9798400711961},
publisher = {Association for Computing Machinery},
address = {New York, NY, USA},
url = {https://doi.org/10.1145/3689031.3696075},
doi = {10.1145/3689031.3696075},
booktitle = {Proceedings of the Twentieth European Conference on Computer Systems},
pages = {1279–1297},
numpages = {19},
location = {Rotterdam, Netherlands},
series = {EuroSys '25}
}

@inproceedings{
wu2024do,
title={Do Language Models Plan Ahead for Future Tokens?},
author={Wilson Wu and John Xavier Morris and Lionel Levine},
booktitle={First Conference on Language Modeling},
year={2024},
url={https://openreview.net/forum?id=BaOAvPUyBO}
}

@misc{lindsey2025biology,
  title={On the Biology of a Large Language Model},
  author={Lindsey, Jack and Gurnee, Wes and Ameisen, Emmanuel and Chen, Brian and Pearce, Adam and Turner, Nicholas L. and Citro, Craig and Abrahams, David and Carter, Shan and Hosmer, Basil and Marcus, Jonathan and Sklar, Michael and Templeton, Adly and Bricken, Trenton and McDougall, Callum and Cunningham, Hoagy and Henighan, Thomas and Jermyn, Adam and Jones, Andy and Persic, Andrew and Qi, Zhenyi and Thompson, T. Ben and Zimmerman, Sam and Rivoire, Kelley and Conerly, Thomas and Olah, Chris and Batson, Joshua},
  howpublished={\url{https://transformer-circuits.pub/2025/attribution-graphs/biology.html}},
  year={2025},
  note={Transformer Circuits Thread}
}

@misc{o3systemcard,
      title={OpenAI o3 and o4-mini System Card}, 
      author={OpenAI},
      year={2025},
      url={https://openai.com/index/o3-o4-mini-system-card/}, 
}

@misc{openai2026chatgpt,
  author       = {{OpenAI}},
  title        = {GPT-5.5 System Card},
  year         = {2026},
  url = {https://openai.com/index/gpt-5-5-system-card/},
}

@inproceedings{NEURIPS2025_5dd3a72b,
 author = {Zhan, Runzhe and Huang, Zhihong and Yang, Xinyi and Chao, Lidia and Yang, Min and Wong, Derek},
 booktitle = {Advances in Neural Information Processing Systems},
 editor = {D. Belgrave and C. Zhang and H. Lin and R. Pascanu and P. Koniusz and M. Ghassemi and N. Chen},
 pages = {64855--64882},
 publisher = {Curran Associates, Inc.},
 title = {Are Large Reasoning Models Good Translation Evaluators? Analysis and Performance Boost},
 url = {https://proceedings.neurips.cc/paper_files/paper/2025/file/5dd3a72bc18a1296ff6070fe4e2be3d0-Paper-Conference.pdf},
 volume = {38},
 year = {2025}
}

@inproceedings{vllm,
author = {Kwon, Woosuk and Li, Zhuohan and Zhuang, Siyuan and Sheng, Ying and Zheng, Lianmin and Yu, Cody Hao and Gonzalez, Joseph and Zhang, Hao and Stoica, Ion},
title = {Efficient Memory Management for Large Language Model Serving with PagedAttention},
year = {2023},
isbn = {9798400702297},
publisher = {Association for Computing Machinery},
address = {New York, NY, USA},
url = {https://doi.org/10.1145/3600006.3613165},
doi = {10.1145/3600006.3613165},
booktitle = {Proceedings of the 29th Symposium on Operating Systems Principles},
pages = {611–626},
numpages = {16},
location = {Koblenz, Germany},
series = {SOSP '23}
}

@inproceedings{luo-etal-2025-hw,
    title = "{HW}-{TSC}{'}s Submissions to the {WMT} 2025 Segment-level Quality Score Prediction Task",
    author = "Luo, Yuanchang  and
      Guo, Jiaxin  and
      Wei, Daimeng  and
      Shang, Hengchao  and
      Li, Zongyao  and
      Rao, Zhiqiang  and
      Yang, Jinlong  and
      Wu, Zhanglin  and
      Chen, Xiaoyu  and
      Yang, Hao",
    editor = "Haddow, Barry  and
      Kocmi, Tom  and
      Koehn, Philipp  and
      Monz, Christof",
    booktitle = "Proceedings of the Tenth Conference on Machine Translation",
    month = nov,
    year = "2025",
    address = "Suzhou, China",
    publisher = "Association for Computational Linguistics",
    url = "https://aclanthology.org/2025.wmt-1.71/",
    doi = "10.18653/v1/2025.wmt-1.71",
    pages = "969--973",
    ISBN = "979-8-89176-341-8"
}

@inproceedings{
ye2025how,
title={How do Transformers Learn Implicit Reasoning?},
author={Jiaran Ye and Zijun Yao and Zhidian Huang and Liangming Pan and Jinxin Liu and Yushi Bai and Amy Xin and Liu Weichuan and Xiaoyin Che and Lei Hou and Juanzi Li},
booktitle={The Thirty-ninth Annual Conference on Neural Information Processing Systems},
year={2025},
url={https://openreview.net/forum?id=19ygs48nOa}
}

\clearpage
\appendix

\section{Data Details}
\label{app:datadetails}

We evaluate on the WMT 2023 test sets for zh-en, en-de, and en-mr, and the WMT 2022 test set for en-ru. We use the training data provided by WMT 2022 and 2023, ensuring no test data leakage. Table \ref{tab:data-statistics} summarizes the number of training and test samples across all language pairs.

\begin{table}[ht]
    \centering
    % \footnotesize
    \begin{tabular}{lcc}
        \toprule
        \textbf{Language Pair} & \textbf{\# Train} & \textbf{\# Test} \\
        \midrule
        zh-en  & 68,182 & 1,664 \\
        en-de  & 47,734 & 1,887 \\
        en-ru  & 26,664 & 511 \\
        en-mr  & 23,329 & 991 \\
        % \midrule
        % \textbf{Total} & \textbf{165,909} & \textbf{5,053} \\
        \bottomrule
    \end{tabular}
    \caption{Training and test data statistics across language pairs}
    \label{tab:data-statistics}
\end{table}

For the UnitSegmentation subtask, we use Qwen3-14B model to generate segmentation results. We then conduct a rule-based unit merge to prevent ground truth error spans from being improperly split across two units. and train Qwen3-4B-Thinking-2507 bachbone with these data. 

In the ErrorDetection subtask, since error-free samples significantly outnumber erroneous ones, we downsample the error-free instances to create a balanced training set. For the SFT stage, we maintain a balanced 1:1 ratio between error-free and erroneous samples for both sentence-level and unit-level ErrorDetection. For RLVR, we empirically set error-free samples to account for 30\% of the overall sentence-level ErrorDetection training data, while still preserving the 1:1 ratio for unit-level ErrorDetection. We find that including an excessive proportion of error-free samples during RLVR training tends to cause the model to collapse into an overly conservative strategy, where it maximizes reward by classifying most samples as error-free.

The remaining subtasks do not involve class imbalance issues, so all available data are retained without additional rebalancing.

\section{RLVR Reward Details}
\label{app:rewardDetails}
This section details the reward design in RLVR training. When the model's output does not include the ``</think>'' token indicating the end of reasoning, or the output JSON format does not meet requirements, or the output length is too short (less than 1000 characters), the reward is set to a punitive -1. Under all other normal circumstances, the reward is computed according to the method described in \ref{sec:synergy}, yielding a value between 0 and 1. For ErrorDetection subtask training samples that do not have ground-truth error spans, the reward is 1 if the model indicates no error spans, and 0 if it indicates any error spans.

This reward design encourages the model to generate code that meets the requirements, even if its capability is insufficient to obtain a normal reward. Since the output format has largely been mastered during the SFT process, this restriction primarily helps the model restore its reasoning ability in the early stages of RLVR and learn to output reasoning chains of valid length.

\section{Training Hyperparameters}
\label{app:trainingDetails}
The SFT training uses a lora rank of 64, a batch size of 64, a learning rate of 1e-4, and a cosine scheduler, running for 2 epochs. 

The RLVR training uses GRPO algorithm, with a rollout group size of 8, a batch size of 128, and a learning rate of 2e-6. The max response length is set to 3000 tokens. The model is trained for up to 6 epochs. To force the model after \textit{NonThinking-SFT} to start explicit reasoning, we put \verb|<think>\nOkay,| in the input.

\section{Evaluation Details}
\label{app:multiple-deocde}
When testing the RIEQE model, we decoded each sample five times to reduce the variability introduced by LLM sampling and to obtain more stable results. The results were aggregated via majority voting. Here, we report the performance of the RIEQE model on the WMT 2023 en–de test set using checkpoints from training epochs 1 to 6, under both single decoding and five-time decoding settings. As shown in Table \ref{tab:multiple_decoding_comparison}, the performance with multiple decodings is more stable.

\begin{table}[htbp]
\centering
\begin{tabular}{ccc}
\toprule
Epochs & Single Decode & Five-time Decode \\
\midrule
1 ep & 0.1905 & 0.2124 \\
2 ep & 0.2395 & 0.2519 \\
3 ep & 0.2169 & 0.2610 \\
4 ep & 0.2381 & 0.2584 \\
5 ep & 0.2688 & 0.2757 \\
6 ep & 0.2567 & 0.2857 \\
\bottomrule
\end{tabular}
\caption{Performance comparison of single decoding vs. five-time decoding across different training epochs (RIEQE model)}
\label{tab:multiple_decoding_comparison}

\end{table}

After two-stage training, the RIEQE model possesses both the ability to provide an answer directly after implicit reasoning and the ability to give a final answer after long explicit reasoning. For the former, we provide an empty reasoning chain \verb|<think>\n\n</think>| in the input and let the model directly output the answer. For the latter, we provide \verb|<think>\nOkay,| in the input and let the model continue reasoning to enforce a long reasoning process. All reasoning processes are conducted using the vLLM \citep{vllm} framework with at most 20000 tokens and the temperature of 1.0.

\section{Human Annotator Consistency}
\label{app:humanAnnotatorConsistency}
The official WMT2025 team provided the officially annotated dataset to another group of human annotators and calculated these annotators' precision, recall, and F1 scores based on the official annotations~\citep{wmt2025qefindings}. Table \ref{tab:human2_high_resource} shows the results for relatively high-resource language pairs (Table 13 in their original paper). The table reveals a low agreement rate among different human annotators, indicating that a significant portion of error spans in fine-grained QE annotation may stem from individual annotator preferences.

\begin{table}[htpb]
\centering
\begin{tabular}{l c c c}
\toprule
\multirow{2}{*}{\textbf{Language Pair}} & \multicolumn{3}{c}{\textbf{Human2}} \\
\cmidrule(lr){2-4}
 & \textbf{P (\%)} & \textbf{R (\%)} & \textbf{F1 (\%)} \\
\midrule
CS-DE\_DE & 30.46 & 41.08 & 34.98 \\
EN-CS\_CZ & 14.40 & 24.86 & 18.24 \\
EN-IT\_IT & 30.52 & 30.62 & 30.57 \\
EN-JA\_JP & 10.61 & 13.93 & 12.04 \\
EN-RU\_RU & 25.28 & 27.56 & 26.37 \\
EN-ZH\_CN & 11.84 & 12.82 & 12.31 \\
\bottomrule
\end{tabular}
\caption{Human2 annotation performance (Micro-F1) for relatively high-resource language pairs.}
\label{tab:human2_high_resource}
\end{table}

\section{Training Discriminability Probes}
\label{app:discriminability}

We train an independent multi-layer perceptron (MLP) probe for each layer (layer-wise) based on the hidden representations. The labels are the model's implicit reasoning output. Each MLP consists of two fully connected layers with ReLU activation and dropout regularization in between. The input dimension is 2560, the hidden dimensions are 256 and 128, and the final output is a scalar logit.

During training, we employ a weighted binary cross-entropy loss (BCE with logits) to mitigate class imbalance. Concretely, the loss for positive samples is scaled by a weight:

\begin{equation}
    w = \frac{N_{\text{neg}}}{N_{\text{pos}}}
\end{equation}

where ($N_{\text{pos}}$) and ($N_{\text{neg}}$) denote the number of positive and negative samples in the training set, respectively. Model parameters are optimized using the Adam optimizer with a learning rate of 2e-5 and a batch size of 256.

We adopt the area under the ROC curve (ROC-AUC) and normalized Average Precision (nAP) as our primary evaluation metrics. ROC-AUC measures the model’s overall ranking capability and is insensitive to class imbalance. In contrast, nAP is designed to enable fair comparison across different class distributions.

Specifically, under different training settings, the proportion of positive samples (i.e., error tokens) varies, leading to different random baselines for Average Precision (AP), which corresponds to the area under the Precision–Recall curve. To eliminate this discrepancy, we normalize AP as follows:

\begin{equation}
    \mathrm{nAP} = \frac{\mathrm{AP} - k}{1 - k}
\end{equation}

where $k$ denotes the proportion of positive samples in the corresponding dataset (i.e., the random baseline AP). This normalization maps a random classifier to 0 and a perfect classifier to 1, thereby ensuring comparability across different data distributions.

During evaluation, for each layer-wise trained classifier, we compute the corresponding ROC-AUC and nAP on a fixed test set and plot the layer-wise performance curves. To estimate the uncertainty of these metrics, we construct 95\% confidence intervals using bootstrap resampling. Specifically, we perform 1000 sampling iterations with replacement on the test set, recompute the evaluation metrics on each resampled subset, and obtain an empirical distribution. The 2.5th and 97.5th percentiles of this distribution are taken as the bounds of the 95\% confidence interval. The error bars in the figures correspond to this interval.

\section{Error Spans Cross Units}
\label{app:errorSpansCrossUnits}
\begin{figure*}
    \centering
    \includegraphics[width=\linewidth]{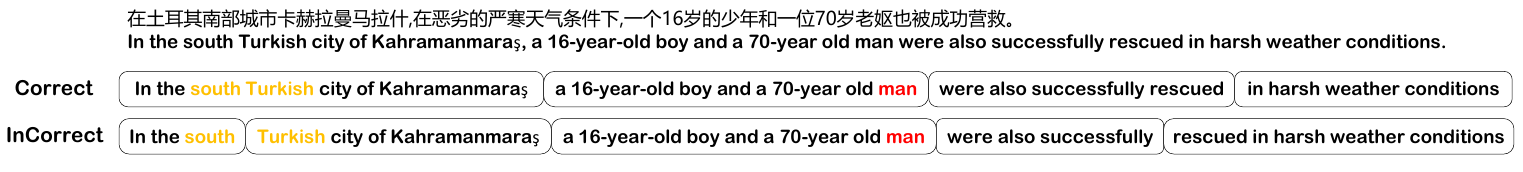}
    \caption{An example of cross-unit error span with task decomposition. The yellow span ``south turkey'' means a MINOR error, which should be ``southern Turkey''; and the red span ``man'' is a MAJOR error, which should be ``woman''.}
    \label{fig:decomposition}
\end{figure*}

In our pipeline, inappropriate segmentation in the first subtask \textbf{UnitSegmentation} at test time may affect the subsequent subtask \textbf{ErrorDetection}. Taking the example in Figure \ref{fig:decomposition}, ``south Turkey'' can be considered as a whole and labeled as a MINOR error. The better phrase could be ``southern Turkish''. If the model splits ``south'' and ``Turkish'' into different units in step 1, then in step 2 it will only see Unit 1: ``... south'' and ``Turkish ...''. In the former unit, the model can identify that ``south'' should be ``southern''; while in the latter, ``Turkish'' appearing alone may no longer be recognized as an error. 

In practice, however, when performing \textbf{UnitSegmentation}, LLMs tend to segment sentences into semantically complete units, which makes it very unlikely for phrases like ``south Turkish'' to be separated. In fact, our model splits the ground truth error span into different units in only about 3\% of the samples in the test sets. Moreover, upon our human inspection, most of these cases are due to overly long annotations, where the ground-truth span may cross multiple semantic segments. And in most of the cases where the annotation is correct, our RIEQE model can accurately identify these overly long error spans by treating consecutive segmented units as errors. We provide a caes study of this phenomena in Appendix \ref{app:casestudy}.

In addition, we experimented with incorporating \textbf{UnitSegmentation} samples during the RLVR stage. The reward only requires that the units segmented by the model are balanced in length and can be restored to the original sentence. This resulted in a word-level MCC of 0.2756 on the WMT en-de-2023 test set, worse than our best performance of 0.2857. We attribute this to the introduction of an additional task during the RL stage, which interferes with the model’s learning.

\section{Performance Across Error Types}
We use the WMT 2024 en-de training set\footnote{\url{https://github.com/google/wmt-mqm-human-evaluation/blob/main/generalMT2024/mqm_generalMT2024_ende.tsv}} to evaluate the model performance on each error type, in order to demonstrate the improvements brought by our training. This training set contains labels for error categories and was not used in our training process. For statistical stability, we only focus on error types that appear more than ten times in the dataset. Table \ref{tab:error-types} presents the results, showing that compared with the backbone model, our model after training achieves significant improvements in both F1 and MCC for most error types, while maintaining a similar level of performance on only a few individual error types. This indicates that the trained model achieves more comprehensive and balanced performance across different categories in the fine-grained QE task.

\begin{table*}[t]
  \centering
  \small
  \setlength{\tabcolsep}{3pt}
  \begin{tabular}{lcccccc}
    \toprule
    & \multicolumn{2}{c}{Qwen3-4B-Thinking-2507} & \multicolumn{2}{c}{RIEQE-NonThinking} & \multicolumn{2}{c}{RIEQE} \\
    \cmidrule(lr){2-3} \cmidrule(lr){4-5} \cmidrule(lr){6-7}
    Error Type & F1 (\%) & MCC & F1 (\%) & MCC & F1 (\%) & MCC \\
    \midrule
    Accuracy/\allowbreak Addition & 9.9 & -0.0099 & 14.6 & 0.1050 & 30.6 & 0.2428 \\
    Accuracy/\allowbreak Creative Reinterpretation & 26.6 & 0.2449 & 25.8 & 0.2378 & 28.6 & 0.2595 \\
    Accuracy/\allowbreak Mistranslation & 39.4 & 0.3566 & 40.5 & 0.3615 & 40.3 & 0.3625 \\
    Accuracy/\allowbreak Source language fragment & 29.5 & 0.2183 & 59.5 & 0.5453 & 61.1 & 0.5640 \\
    Fluency/\allowbreak Grammar & 18.2 & 0.1240 & 28.9 & 0.2627 & 27.6 & 0.2573 \\
    Fluency/\allowbreak Inconsistency & 17.5 & 0.1136 & 19.0 & 0.1531 & 22.1 & 0.1747 \\
    Fluency/\allowbreak Punctuation & 8.9 & 0.0499 & 7.3 & 0.0424 & 9.8 & 0.0492 \\
    Fluency/\allowbreak Register & 14.8 & 0.0759 & 22.0 & 0.2271 & 21.1 & 0.2214 \\
    Fluency/\allowbreak Spelling & 17.0 & 0.1287 & 25.0 & 0.2200 & 25.1 & 0.2278 \\
    Fluency/\allowbreak Text-Breaking & 4.9 & 0.0252 & 13.3 & 0.1301 & 13.3 & 0.1301 \\
    Non-translation! & 21.4 & 0.0858 & 72.7 & 0.2408 & 76.8 & 0.2684 \\
    Other & 27.6 & 0.2299 & 62.8 & 0.6227 & 58.6 & 0.5886 \\
    Style/\allowbreak Bad sentence structure & 8.7 & -0.0005 & 12.9 & 0.1366 & 11.0 & 0.1642 \\
    Style/\allowbreak Unnatural or awkward & 19.8 & 0.1422 & 18.6 & 0.1494 & 18.7 & 0.1532 \\
    Terminology/\allowbreak Inappropriate for context & 23.6 & 0.2171 & 28.9 & 0.2694 & 33.1 & 0.3166 \\
    \midrule
    \textbf{Overall (Macro-Avg)} & \textbf{19.2} & \textbf{0.1334} & \textbf{30.1} & \textbf{0.2469} & \textbf{31.9} & \textbf{0.2654} \\
    \bottomrule
  \end{tabular}
  \caption{Per-error-type token-level evaluation results on the WMT24 en-de MQM train set. We report token-level F1 and Matthews Correlation Coefficient (MCC) for each error type.}
  \label{tab:error-types}
\end{table*}

\section{Case Study}
Figure 
\label{app:casestudy}
\ref{fig:rieqeexample} presents a typical example from zh-en language pair. In this section, we first demonstrate an example from en-de to illustrate how surface fluency can mislead LLM judgments, and how the model trained under the RIEQE framework can accurately detect the error. We then present a second example to show that our model can also effectively handle the excessively long error span annotations across multiple units discussed in Appendix~\ref{app:errorSpansCrossUnits}.

\paragraph{The en-de example.}
Consider the following QE case:
\begin{quote}
\textbf{src:} ``Because if the paper is worth publishing.''\\ \textbf{mt:} ``Denn wenn die \underline{Zeitung} es wert ist, veröffentlicht zu werden.''
\end{quote}
The error lies in the mistranslation of the word paper. In the source sentence, ``paper'' clearly refers to an academic paper. The German translation, however, renders it as ``Zeitung'' (newspaper), which is semantically incorrect. Compounding this issue, the mistranslated word happens to fit fluently into the syntactic structure of the target sentence. This surface fluency makes the semantic error particularly subtle and easy to overlook. As a result, an LLM without RIEQE training may fail to detect it, whereas the model trained under the RIEQE framework is able to identify the problem.

\paragraph{The overly long annotation example.} Next, consider a case with an excessively long error span:
\begin{quote}
\textbf{src:} ``Throwing everything you've got over the fence in response to GPT is not it.''\\ \textbf{mt:} ``\underline{Alles, was du hast, über den Zaun} \underline{zu werfen, als Reaktion auf GPT, ist es} \underline{nicht.}''
\end{quote}
In this example, the translation is a literal translation and has a chaotic structure. Therefore, the entire sentence is annotated as a translation error, forming a single error span. The RIEQE model decomposes it into four units and then marks the content in all four units as erroneous. This demonstrates that excessively long error spans do not impair the model's judgment capability at inference time.

\section{Inference Time of RIEQE}
\label{app:inferenceTime}
As an LRM-based QE method, a key concern is the inference efficiency of our approach, which affects both its potential as a machine translation reward model and the efficiency of practical applications involving large-batch quality inspection of translations. Using the Qwen3-4B-Thinking-2507 as the backbone for our RIEQE in this paper, we tested the inference latency on the en–de test set (1,887 samples) using vLLM 0.8.5.post1 on an RTX 4090 (48GB). The actual inference costs are in Table \ref{tab:inference_latency}.

\begin{table*}[htbp]
\centering
\caption{Inference latency of RIEQE on the en--de test set (1,887 samples $\times$ 5 decodings, covering three subtasks).}
\label{tab:inference_latency}
\begin{tabular}{l c c c}
\hline
\textbf{Mode} & \textbf{GPU} & \textbf{Total time cost} & \textbf{Time per sample} \\
\hline
NonThinking & 1$\times$ RTX 4090 & 420s (115s + 250s + 55s) & 0.22s \\
Thinking & 2$\times$ RTX 4090 & 1459s (598s + 567s + 294s) & 0.77s \\
\hline
\end{tabular}
\end{table*}

As can be seen, NonThinking mode requires only 0.22s per sample on average, making it a viable choice for the RLVR reward model. Thinking mode (using 2 GPUs) averages 0.77s per sample, which is perfectly acceptable in settings with sufficient compute, while delivering the best QE performance.

It should be noted that the above times were measured under a high-throughput batch scenario of 1,887 samples $\times$ 5 decodings, where vLLM's continuous batching and PagedAttention fully realize their acceleration advantages. In single-sample or small-batch request scenarios, per-sample latency may be higher than the figures reported here. We therefore believe that RIEQE has potential as a reward model when the batch size is reasonably large.

\section{Prompts}
\label{app:prompts}

In this section, we present the prompts used for subtask training in our method pipeline, as well as the prompt used for the wholetask setting in \S\ref{sec:wholetask}. 

Figures \ref{fig:prompt1}, \ref{fig:prompt2}, \ref{fig:prompt3} show the prompts for the three subtasks UnitSegmentation, ErrorDetection and SeverityClassification, respectively. Figures \ref{fig:prompt2-oneUnit} and \ref{fig:prompt3-oneSpan} show the prompts for the subtasks ErrorDetection and SeverityClassification with only one unit or error span input. Figure \ref{fig:prompt-wholetask} shows the prompt for the WholeTask setting.

\begin{figure*}[t]
\centering
\fbox{
\begin{minipage}{0.95\textwidth}
\small
Divide the sentence into **small semantic units** (around 5 words each).

\vspace{1em}
Each unit should represent a relatively complete piece of meaning.

\vspace{1em}
All units must be exact spans from the original sentence, and when concatenated in order, they should reconstruct the original sentence without any modification.

\vspace{1em}
Output in JSON format:

\{

\quad "Unit1": <SemanticUnit1>,

\quad "Unit2": <SemanticUnit2>,

\quad ...

\}

\vspace{1em}
Sentence:

\{tgt\}
\end{minipage}
}
\caption{Prompt for subtask 1: Unit Segmentation.}
\label{fig:prompt1}
\end{figure*}

\begin{figure*}[t]
\centering
\fbox{
\begin{minipage}{0.95\textwidth}
\small
Given the following translation pair:

\vspace{1em}
Source Sentence:

\{src\}

Target Sentence:

\{tgt\}

\vspace{1em}
If we segment the target sentence into several semantic units:

\{units\}

\vspace{1em}
Inspect each semantic unit **in order from left to right**.

\vspace{1em}
For each unit, check the following dimensions (no re-evaluation or backtracking):

1. **Terminology consistency** — Are key terms translated consistently and correctly? 

2. **Faithfulness / Accuracy** — Does the meaning match the source sentence precisely? 

\quad - Detect any mistranslation, omission, or addition.  

3. **Language naturalness and style** — Is the expression idiomatic, fluent, and appropriate in register?

\vspace{1em}
For each semantic unit containing an error, identify the error span within its corresponding translation.

- The `span` must be an **exact substring** from the target sentence.  

- The span should cover the **entire erroneous phrase** that conveys the incorrect meaning, not just an individual incorrect token.  

\quad - Include all words that together form the wrong meaning.  

\quad - Do **not** include unrelated correct words.  

- If there are multiple errors, list them **in the order they appear** in the target sentence.

\vspace{1em}
Output in JSON format:

\{

\quad "Unit1": [

\quad \quad "<error span 1>",
        
\quad \quad "<error span 2>",

\quad \quad ...

\quad],

\quad...

\}
\end{minipage}
}
\caption{Prompt for subtask 2: Error Detection.}
\label{fig:prompt2}
\end{figure*}

\begin{figure*}[t]
\centering
\fbox{
\begin{minipage}{0.95\textwidth}
\small
Given the following translation pair:

\vspace{1em}
Source Sentence:

\{src\}

Target Sentence:

\{tgt\}

\vspace{1em}
If we segment the target sentence into several semantic units:

\{units\}

\vspace{1em}
And we only examine this semantic unit:

\{unit\}

\vspace{1em}
For this unit, check the following dimensions (no re-evaluation or backtracking):

1. **Terminology consistency** — Are key terms translated consistently and correctly? 

2. **Faithfulness / Accuracy** — Does the meaning match the source sentence precisely? 

\quad - Detect any mistranslation, omission, or addition.  

3. **Language naturalness and style** — Is the expression idiomatic, fluent, and appropriate in register?

\vspace{1em}
For each semantic unit containing an error, identify the error span within its corresponding translation.

- The `span` must be an **exact substring** from the target sentence.  

- The span should cover the **entire erroneous phrase** that conveys the incorrect meaning, not just an individual incorrect token.  

\quad - Include all words that together form the wrong meaning.  

\quad - Do **not** include unrelated correct words.  

- If there are multiple errors, list them **in the order they appear** in the target sentence.

\vspace{1em}
Output in JSON format:

\{

\quad "errors": [

\quad \quad "<error span 1>",
        
\quad \quad "<error span 2>",

\quad \quad ...

\quad]

\}
\end{minipage}
}
\caption{Prompt for subtask 2: Error Detection, with only one unit input.}
\label{fig:prompt2-oneUnit}
\end{figure*}

\begin{figure*}[t]
\centering
\fbox{
\begin{minipage}{0.95\textwidth}
\small
Given the following translation pair:

\vspace{1em}
Source Sentence:

\{src\}

Target Sentence:

\{tgt\}

\vspace{1em}
If we segment the target sentence into several semantic units:

\{units\}

\vspace{1em}
And we know all the error spans:

\{spans\}

\vspace{1em}
Assign each error span with an error severity MINOR or MAJOR:

- **MINOR** — Minor grammatical, stylistic, or fluency issues that do not significantly alter meaning.  

- **MAJOR** — Errors that change, distort, or obscure the meaning, including mistranslation, omission, or addition.

\vspace{1em}
Output in JSON format:

\{

\quad "errors": [

\quad \quad \{
    
\quad \quad \quad "span": "<error span>",
        
\quad \quad \quad "severity": "<MINOR or MAJOR>"
        
\quad \quad \},
    
\quad \quad ...
    
\quad ]
    
\}
\end{minipage}
}
\caption{Prompt for subtask 3: Severity Classification.}
\label{fig:prompt3}
\end{figure*}

\begin{figure*}[t]
\centering
\fbox{
\begin{minipage}{0.95\textwidth}
\small
Given the following translation pair:

\vspace{1em}
Source Sentence:

\{src\}

Target Sentence:

\{tgt\}

\vspace{1em}
If we segment the target sentence into several semantic units:

\{units\}

\vspace{1em}
And we know all the error spans:

\{spans\}

\vspace{1em}
And we only examine the error span:

\{span\}

\vspace{1em}
For this span, assign an error severity MINOR or MAJOR:

- **MINOR** — Minor grammatical, stylistic, or fluency issues that do not significantly alter meaning.  

- **MAJOR** — Errors that change, distort, or obscure the meaning, including mistranslation, omission, or addition.

\vspace{1em}
Output in JSON format:

\{

\quad "errors": [

\quad \quad \{

\quad \quad \quad "span": "<error span>",

\quad \quad \quad "severity": "<MINOR or MAJOR>"
    
\quad \quad \}
    
\quad ]
    
\}
\end{minipage}
}
\caption{Prompt for subtask 3: Severity Classification, with only one error span input.}
\label{fig:prompt3-oneSpan}
\end{figure*}

\begin{figure*}[t]
\centering
\fbox{
\begin{minipage}{0.95\textwidth}
\small
You are doing a translation quality estimation task.

\vspace{1em}
Given the following translation pair:

\vspace{1em}
Source Sentence:

\{src\}

Target Sentence:

\{tgt\}

\vspace{1em}
To find translation errors, check the following dimensions (no re-evaluation or backtracking):

1. **Terminology consistency** — Are key terms translated consistently and correctly? 

2. **Faithfulness / Accuracy** — Does the meaning match the source sentence precisely?  

\quad - Detect any mistranslation, omission, or addition.  

3. **Language naturalness and style** — Is the expression idiomatic, fluent, and appropriate in register?

\vspace{1em}
For each error, identify the error span within its corresponding translation.

- The `span` must be an **exact substring** from the target sentence.  

- The span should cover the **entire erroneous phrase** that conveys the incorrect meaning, not just an individual incorrect token.  

\quad - Include all words that together form the wrong meaning.  
\quad - Do **not** include unrelated correct words.  

- If there are multiple errors, list them **in the order they appear** in the target sentence.

\vspace{1em}
For each error span, assign an error severity MINOR or MAJOR:

- **MINOR** — Minor grammatical, stylistic, or fluency issues that do not significantly alter meaning.  

- **MAJOR** — Errors that change, distort, or obscure the meaning, including mistranslation, omission, or addition.

\vspace{1em}
Output in JSON format:

\{

\quad "errors": [

\quad \quad \{

\quad \quad \quad "span": "<error span>",

\quad \quad \quad "severity": "<MINOR or MAJOR>"
    
\quad \quad \},

\quad \quad ...
    
\quad ]
    
\}
\end{minipage}
}
\caption{Prompt for the WholeTask setting.}
\label{fig:prompt-wholetask}
\end{figure*}

\end{document}